\documentclass[runningheads]{llncs}

\usepackage{eccv}

\usepackage{eccvabbrv}

\usepackage{graphicx}
\usepackage{booktabs}
\usepackage[many]{tcolorbox}
\usepackage[accsupp]{axessibility}  

\usepackage{hyperref}

\usepackage{orcidlink}
\usepackage{booktabs}
\usepackage{multirow}
\usepackage{graphicx}
\usepackage{xcolor}
\usepackage{colortbl}
\usepackage{algorithm}
\usepackage{algorithmic}

\definecolor{bestred}{HTML}{C0392B} 
\definecolor{bestblue}{HTML}{2980B9} 
\definecolor{graybg}{HTML}{F2F3F4}

\newcommand{\best}[1]{\textbf{\textcolor{bestred}{#1}}}
\newcommand{\second}[1]{\textbf{\textcolor{bestblue}{#1}}}
\newcommand{\sectionrow}[1]{\multicolumn{9}{l}{\cellcolor{graybg}\textbf{\textit{#1}}}}
\usepackage{colortbl}

\definecolor{bestred}{HTML}{C0392B}
\definecolor{bestblue}{HTML}{2980B9}
\definecolor{graybg}{HTML}{F2F3F4}

\begin{document}

\title{Agentic Flow Steering and Parallel Rollout Search for Spatially Grounded Text-to-Image Generation} 

\titlerunning{AFS-Search}


\author{Ping Chen\inst{1}\orcidlink{0009-0006-8039-2746} \and
Daoxuan Zhang\inst{1}\orcidlink{0009-0004-9622-0565} \and
Xiangming Wang\inst{1}\orcidlink{0009-0009-3941-3447} \and
Yungeng Liu\inst{1}\orcidlink{0009-0007-0685-5949} \and
Haijin Zeng\inst{1}\orcidlink{0000-0003-0398-3316}$^{*}$ \and
Yongyong Chen\inst{1}\orcidlink{0000-0003-1970-1993}\thanks{Corresponding authors.}
}


\authorrunning{P. Chen et al.}

\institute{
Harbin Institute of Technology, Shenzhen, China\\
\email{zenghj@hit.edu.cn}\\
\email{cyy2020@hit.edu.cn}
}

\maketitle

\begin{figure*}[htbp]
  \centering
  \includegraphics[width=1\textwidth]{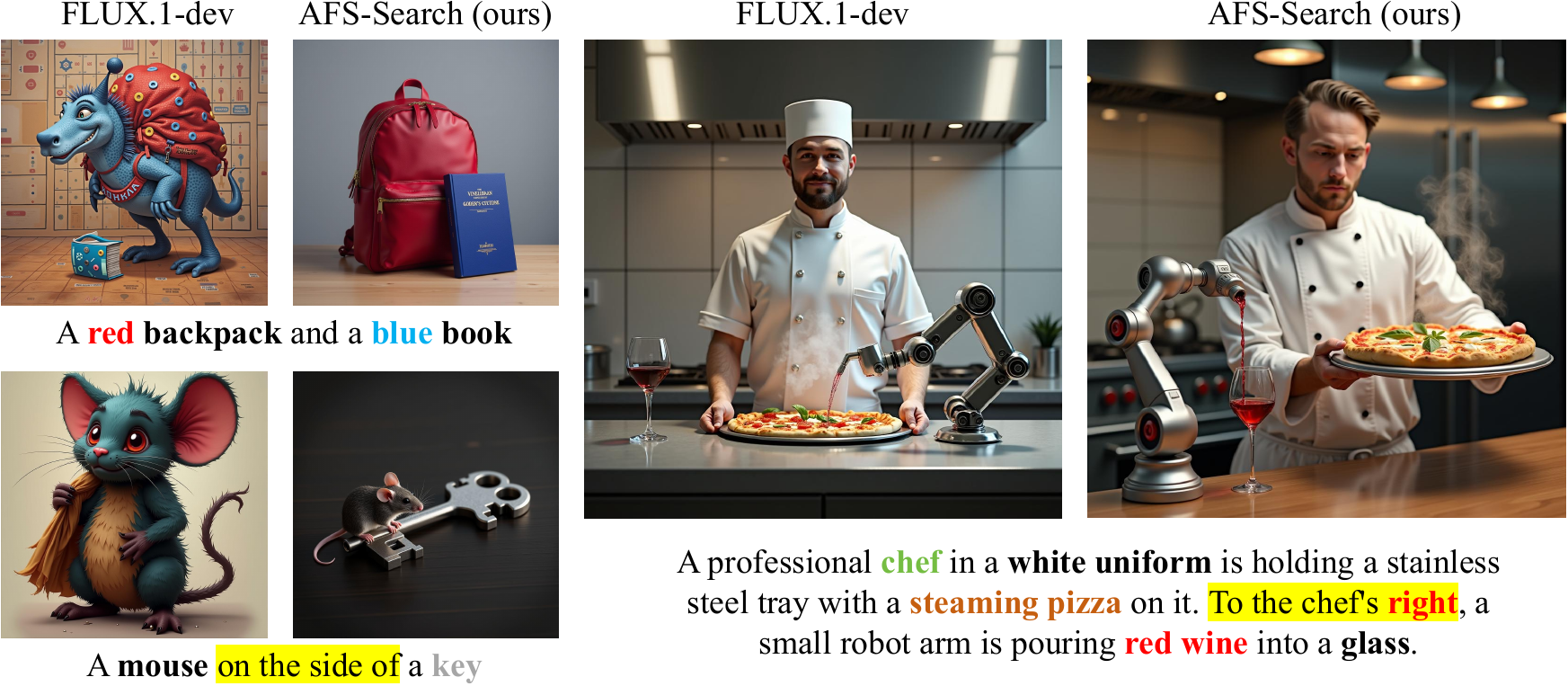} 
  \caption{From a visual perspective, our AFS-Search provides a closed-loop generation paradigm to achieve precise spatial grounding generation.}
  \label{fig:motivation}
\end{figure*}

\begin{abstract}
 Precise Text-to-Image (T2I) generation has achieved great success but is hindered by the limited relational reasoning of static text encoders and the error accumulation in open-loop sampling. Without real-time feedback, initial semantic ambiguities during the Ordinary Differential Equation trajectory inevitably escalate into stochastic deviations from spatial constraints. To bridge this gap, we introduce AFS-Search (\textbf{A}gentic \textbf{F}low \textbf{S}teering and Parallel Rollout \textbf{Search}), a training-free closed-loop framework built upon FLUX.1-dev. AFS-Search incorporates a training-free closed-loop parallel rollout search and flow steering mechanism, which leverages a Vision-Language Model (VLM) as a semantic critic to diagnose intermediate latents and dynamically steer the velocity field via precise spatial grounding. Complementarily, we formulate T2I generation as a sequential decision-making process, exploring multiple trajectories through lookahead simulations and selecting the optimal path based on VLM-guided rewards. Further, we provide AFS-Search-Pro for higher performance and AFS-Search-Fast for quicker generation. Experimental results show that our AFS-Search-Pro greatly boosts the performance of the original FLUX.1-dev, achieving state-of-the-art results across three different benchmarks. Meanwhile, AFS-Search-Fast also significantly enhances performance while maintaining fast generation speed.
  \keywords{T2I Generation \and Vision-Language Model \and AFS-Search}
\end{abstract}

\section{Introduction}
\label{sec:intro}

The field of Text-to-Image (T2I) generation has witnessed a paradigm shift with the emergence of Diffusion Models (DMs)~\cite{DBLP:conf/nips/HoJA20, DBLP:conf/cvpr/RombachBLEO22, DBLP:conf/iclr/SongME21} and Flow Matching (FM)~\cite{DBLP:conf/iclr/LipmanCBNL23, DBLP:journals/tmlr/LiSFH25} architectures. Models such as SDXL~\cite{DBLP:conf/iclr/PodellELBDMPR24}, FLUX.1~\cite{labs2025flux1kontextflowmatching}, and Qwen-Image~\cite{DBLP:journals/corr/abs-2508-02324} have demonstrated an unparalleled ability to synthesize high-fidelity images that capture intricate artistic styles and textures. By leveraging large-scale pre-trained text encoders such as T5~\cite{DBLP:journals/jmlr/RaffelSRLNMZLL20}, CLIP~\cite{DBLP:conf/icml/RadfordKHRGASAM21}, these models have moved beyond simple object depiction toward generating complex scenes from natural language descriptions. Recently, VLM-based frameworks such as RPG~\cite{DBLP:conf/icml/0006YMXE024}, SILMM~\cite{DBLP:conf/cvpr/QuL00LNC25}, AgentComp~\cite{DBLP:journals/corr/abs-2512-09081} encounter, showing great potential to further complete T2I generation via VLM perception and even agentic actions.

Despite these impressive strides, achieving precise spatial grounding and relational reasoning remains a persistent challenge. As shown in Fig.~\ref{fig:teaser}, we identify two primary bottlenecks in conventional T2I pipelines: (1) Static text encoders often exhibit an expressive bottleneck when processing complex relational semantics. Specifically, they struggle to distinguish detailed spatial instructions, which results in underspecified semantic embeddings that fail to capture the nuanced spatial relationships required for accurate image synthesis. (2) Traditional models, as well as recent agent-guided frameworks such as RPG~\cite{DBLP:conf/icml/0006YMXE024}, Layout-Guidance~\cite{DBLP:conf/icassp/SongLLSZCYC25}, AgentComp~\cite{DBLP:journals/corr/abs-2512-09081} follow an open-loop sampling paradigm. While existing agent-based frameworks enhance spatial control through pre-generation planning of the model, they solve an Ordinary Differential Equation (ODE) along a pre-defined trajectory without any internal feedback. Consequently, even minor semantic ambiguities in the initial phase are irreversibly amplified through the discrete integration steps, ultimately leading to stochastic deviations where the final output fails to satisfy the original spatial constraints.

To bridge this gap, we introduce AFS-Search, a training-free \textbf{closed-loop} framework designed to transform T2I generation from a passive sampling process into an active, decision-making procedure. Our core insight is that a generative model should iteratively assess and adjust its generation process rather than producing outputs in a one-shot manner. By integrating a Vision-Language Model (VLM) as a Semantic Critic, our framework enables the system to perceive intermediate generation states and rectify potential errors in real-time. Specifically, we propose \textbf{Agentic Flow Steering (AFS)}  a novel steering mechanism that diagnoses semantic drifts at critical timestamps and dynamically steers the velocity field of the flow model via precise spatial grounding leveraging SAM3~\cite{DBLP:journals/corr/abs-2511-16719}. 

Additionally, going beyond simple correction, our AFS-Search framework incorporates Parallel Rollout Search (PRS) to effectively navigate the complex latent landscape. At key bifurcation points, the agent performs lookahead simulations by exploring multiple potential trajectories, including \textbf{corrective steering and random exploration.} By evaluating these branches through VLM-guided rewards, the model selects the optimal path that maximizes alignment with the user's intent. This strategy effectively leverages test-time computation to overcome the inherent randomness of the diffusion process, thereby ensuring robust performance. Further, we provide AFS-Search-Pro for higher performance and AFS-Search-Fast for quicker generation. The main contributions are as follows:

\begin{figure*}[t]
  \centering
  \includegraphics[width=1\textwidth]{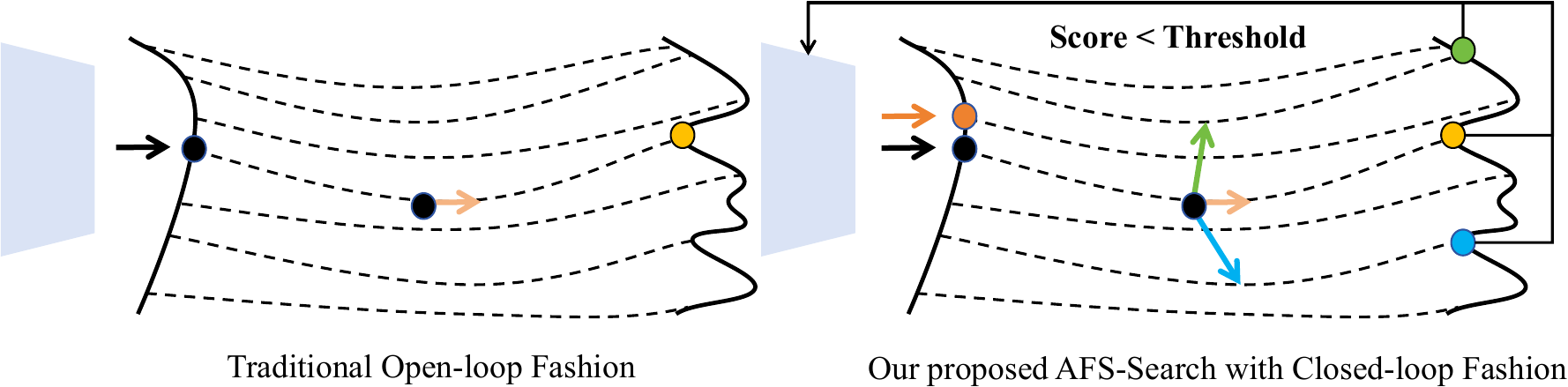} 
  \caption{\textbf{Motivation of our AFS-Search.} Open-loop generation follows a fixed, feed-forward sampling trajectory without intermediate feedback or correction while closed-loop generation introduces real-time visual feedback.}
  \label{fig:teaser}
\end{figure*}

\begin{itemize}
    \item We propose AFS-Search, a training-free framework that reformulates T2I generation as as a closed-loop decision-making process instead of a passive sampling task. We provide two versions of AFS-Search for real-time computation and efficiency.
    \item We introduce Agentic Flow Steering mechanism that dynamically rectifies the ODE trajectory via VLM feedback and spatial anchoring via SAM3.
    \item We implement a Parallel Rollout Search strategy using lookahead simulations to effectively resolve spatial and semantic conflicts during sampling.
    \item Experimental results show that our AFS-Search-Pro achieves state-of-the-art results across three benchmarks, while AFS-Search-Fast also significantly enhances performance while meeting the requirements for fast generation.
\end{itemize}

\section{Related Work}

\subsection{Compositional Text-to-Image Generation}
T2I generation has achieved remarkable progress, evolving from early GAN-based approaches~\cite{DBLP:conf/icml/ReedAYLSL16, DBLP:conf/iccv/ZhangXL17, DBLP:conf/cvpr/XuZHZGH018} to the current state-of-the-art Diffusion Models~\cite{DBLP:conf/cvpr/RombachBLEO22, DBLP:conf/nips/SahariaCSLWDGLA22} and Flow Matching architectures~\cite{DBLP:conf/iclr/LipmanCBNL23, DBLP:journals/tmlr/LiSFH25, labs2025flux1kontextflowmatching}. Despite their ability to synthesize high-quality textures, standard models often struggle with \textit{compositional generation}. Specifically, correctly binding attributes to objects and strictly following spatial instructions. This limitation stems from the cross-attention mechanism's tendency to mix semantic information across different spatial regions, leading to attribute leakage or catastrophic neglect of objects.

Existing solutions can be broadly categorized into training-based and training-free approaches. Training-based methods~\cite{DBLP:conf/iccv/ZhangRA23, DBLP:conf/cvpr/RuizLJPRA23} fine-tune the backbone or introduce additional adapters to enforce spatial constraints. However, in the era of billion-parameter foundation models trained on massive internet-scale datasets, fine-tuning on limited domain-specific data often yields diminishing returns. More critically, which causes the model to lose its open‑world generalizability. 

Consequently, \textbf{training-free} methods have gained prominence. Early works utilized attention manipulation~\cite{DBLP:conf/iclr/HertzMTAPC23, DBLP:journals/tog/CheferAVWC23} to re-weight or mask cross-attention maps. While effective for simple layouts, these heuristic-based methods lack high-level reasoning and often fail in complex scenarios requiring logical planning. Distinct from these, our \textbf{AFS-Search} adopts a test-time search paradigm. We argue that pre-trained models already possess the necessary visual priors; the challenge lies not in learning new features, but in \textit{navigating} the latent space to locate the correct composition without altering model weights.

\subsection{Vision Language Models and Agentic Frameworks}
The rapid evolution of Large Multimodal Models (LMMs)~\cite{DBLP:journals/corr/abs-2303-08774, DBLP:journals/corr/abs-2412-19437} has catalyzed a new wave of Agentic generation frameworks that leverage the reasoning capabilities of LLMs/VLMs to control T2I synthesis. Early works focused on prompt enhancement and layout planning. For instance, DALL-E 3~\cite{DBLP:journals/corr/abs-2310-07653} and Promptist~\cite{promptist} employ LLMs to rewrite user queries into descriptive captions. Moving beyond text, LayoutGPT~\cite{DBLP:conf/nips/FengZFJAHBWW23} and VPGen~\cite{DBLP:conf/nips/0001ZB23} utilize LLMs to generate intermediate scene layouts to guide diffusion models. While effective for initial grounding, these methods operate in an open-loop manner, lacking mechanisms to verify if the generated layout is actually respected.

To address this, recent research has pivoted towards feedback-driven iterative frameworks. Idea2Img~\cite{DBLP:journals/corr/abs-2310-08541} introduces a multi-turn dialogue where an LMM iteratively revises prompts based on generated drafts. RPG~\cite{DBLP:conf/icml/0006YMXE024} proposes a chain-of-thought planning strategy to decompose complex prompts into sub-regions. AgentComp~\cite{DBLP:journals/corr/abs-2512-09081} integrates multiple agent roles to iteratively refine compositional details via an external feedback loop. However, a common limitation of these state-of-the-art agents is their reliance on an external loop paradigm: they treat the generative model as a black box, rectifying errors solely by modifying the textual input and triggering a full re-generation. This trial-and-error process is computationally inefficient and often struggles to correct fine-grained local attributes without altering the global structure.

In contrast, our approach shifts the paradigm from external re-prompting to internal state intervention. Instead of discarding the entire image upon failure, AFS-Search intervenes directly within the flow-matching trajectory. By performing \textit{Parallel Rollout Search} on the intermediate latents, we achieve precise, localized corrections without the computational overhead of iterative re-generation.

\section{Method}
\label{sec:method}

\subsection{Overview}
\begin{figure*}[t]
  \centering
  \includegraphics[width=\textwidth]{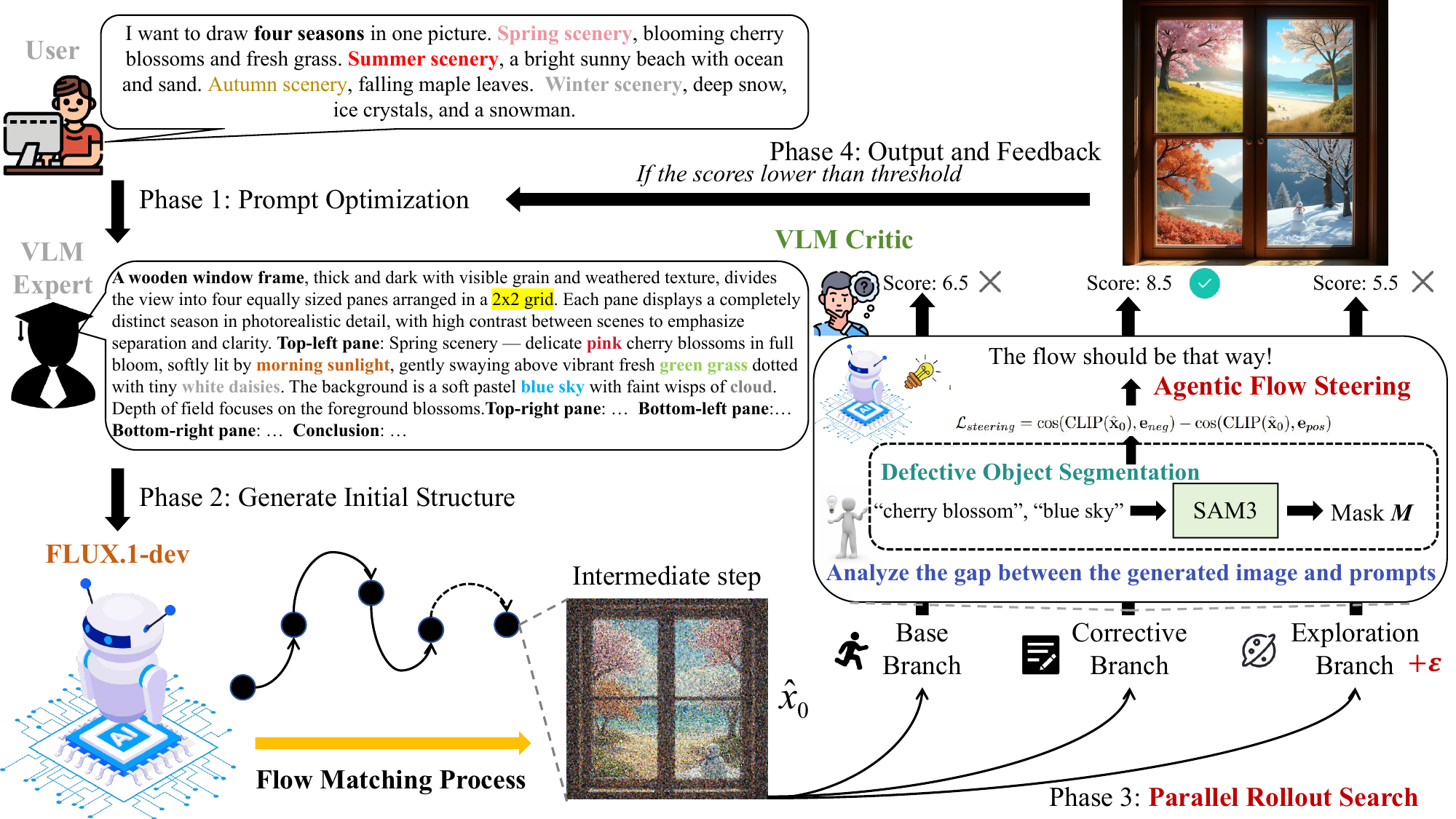} 
  \caption{\textbf{The framework of AFS-Search.} The pipeline operates in four phases: 
  \textbf{(1) Prompt Optimization.} A VLM rewrites the user prompt into an explicit instruction. 
  \textbf{(2) Generating Initial Structure.} The FLUX.1-dev model generates an intermediate state up to a bifurcation point. 
  \textbf{(3) Parallel Rollout Search.} A VLM Critic diagnoses the intermediate state to guide search. The system explores three branches: a \textit{Baseline Branch}, an \textit{Exploration Branch}, and a \textit{Corrective Branch} by AFS. 
  \textbf{(4) Output \& Feedback:} The optimal trajectory is selected based on VLM scores. If the scores are lower than threshold, a global redesign loop is triggered.}
  \label{fig:overview}
\end{figure*}

As illustrated in Fig.~\ref{fig:overview}, \textbf{AFS-Search} reformulates text-to-image generation as a closed-loop decision-making process comprising four integrated phases: 
(1) \textbf{Prompt Optimization}, where a VLM rewrites abstract user inputs into detailed, spatially explicit instructions to minimize initial ambiguity; 
(2) \textbf{Initial Structure Generation}, where the FLUX.1-dev~\cite{labs2025flux1kontextflowmatching} model synthesizes latents up to a critical bifurcation point to establish a malleable global layout; 
(3) \textbf{Parallel Rollout Search}, the core phase where a VLM Critic diagnoses intermediate defects to guide lookahead simulations across three branches: \textit{Base}, \textit{Exploration}, and \textit{Corrective}, which employs SAM3-based Agentic Flow Steering via contrastive guidance and ultimately selects the optimal trajectory based on reward scores; and 
(4) \textbf{Global Feedback}, which concludes the generation and triggers a re-design loop if the final output score falls below a safety threshold.

\subsection{Preliminaries: Flow Matching and Open-Loop Limitation}
\label{subsec:preliminaries}

Based on FLUX.1 dev~\cite{labs2025flux1kontextflowmatching}, we build our method upon the Flow Matching paradigm, which models the generation process as a Continuous Normalizing Flow (CNF). Given a data distribution $q(x_1)$ and a prior distribution $p(x_0) = \mathcal{N}(x_0; 0, I)$, the flow is defined by a time-dependent vector field $v_t(x)$. The generation process involves solving the following ODE:
\begin{equation}
    dx_t = v_\theta(x_t, t, y) dt,
\end{equation}
where $v_\theta$ is a neural network parameterized by $\theta$ and conditioned on text embedding $y$. Standard sampling integrates this ODE from $t=0$ to $t=1$. However, this open-loop integration is prone to error accumulation. Without feedback, any semantic misalignment in intermediate steps propagates to the final output.

\subsection{Parallel Rollout Search (PRS)}
\label{subsec:PRS}

\begin{figure*}[t]
  \centering
  \includegraphics[width=\textwidth]{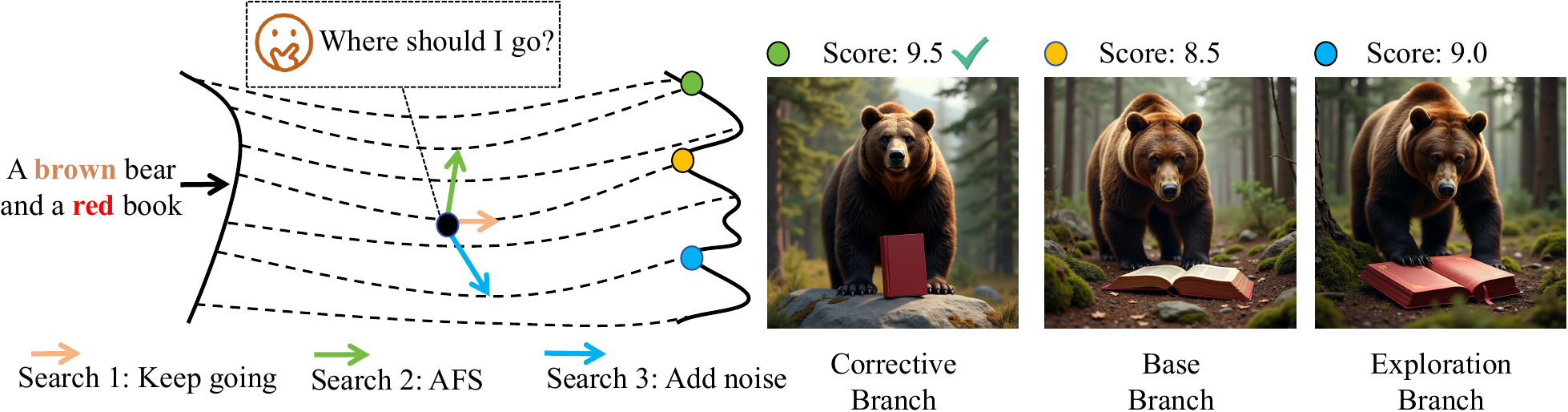} 
  \caption{\textbf{Motivation for Parallel Rollout Search.} While the standard open-loop trajectory (Base Branch) yields a sub-optimal alignment (Score: 8.5), our search mechanism actively explores alternative futures. By comparing the Corrective Branch (guided by AFS) and Exploration Branch against the baseline, the agent identifies and selects the optimal trajectory (Score: 9.5) that best matches the prompt.}
  \label{fig:ifs}
\end{figure*}

\textbf{Motivation.}
Standard open-loop sampling suffers from stochastic failure, where early semantic errors accumulate irreversibly. To address this, as shown in Fig.~\ref{fig:ifs}, we reformulate generation as a \textbf{navigable decision-making process} rather than a fixed probabilistic trajectory. Crucially, we adopt a \textbf{training-free} strategy. Instead of fine-tuning, which risks catastrophic forgetting of the backbone's open-world knowledge, we leverage \textbf{test-time computation} to explore the latent space. This approach unlocks the pre-trained model's inherent capability to follow complex spatial instructions without altering PRS parameters.

\paragraph{\textbf{Prompt Optimization.}}
The process begins with a \textbf{VLM Prompt Optimizer}. Given a user prompt $y_{raw}$, a VLM rewrites it into a comprehensive instruction $y_{refined}$ with defined logic constraints (e.g., object counts and precise colors). This ensures that constraints are explicitly defined before the denoising process begins. \textbf{The agent's prompt is provided in Supplyment~\ref{sec:Full Prompt Templates}.}

\paragraph{\textbf{Latent Space Search Tree.}}
We perform standard denoising until a critical \textit{decision point} $t_{split}$ such as $60\%$ of total steps. At this state $x_{t_{split}}$, the VLM acts as a supervisor to diagnose defects such as object count errors, color mismatches, or spatial deviations. Based on this diagnosis, we construct a search tree with action space $\mathcal{A} = \{a_{base}, a_{steer}, a_{explore}\}$. Specifically:

\noindent \textbf{Baseline Branch ($a_{base}$)} Continues the basic trajectory without intervention. 

\noindent \textbf{Corrective Branch ($a_{steer}$)} activates the \textit{Agentic Flow Steering} module (see Sec.~\ref{subsec:afs}) to rectify specific semantic defects. 

\noindent \textbf{Exploration Branch ($a_{explore}$)} introduces stochastic perturbations to escape potential local semantic optima. By injecting controlled Gaussian noise $\epsilon \sim \mathcal{N}(0, \sigma^2 \mathbf{I})$ into the latent state $x_{t_{split}}$, this branch forces the ODE solver to diverge from the current deterministic path, allowing the model to re-sample alternative global layouts or object poses while preserving the broad context.

\begin{figure*}[t]
  \centering
  \includegraphics[width=1\textwidth]{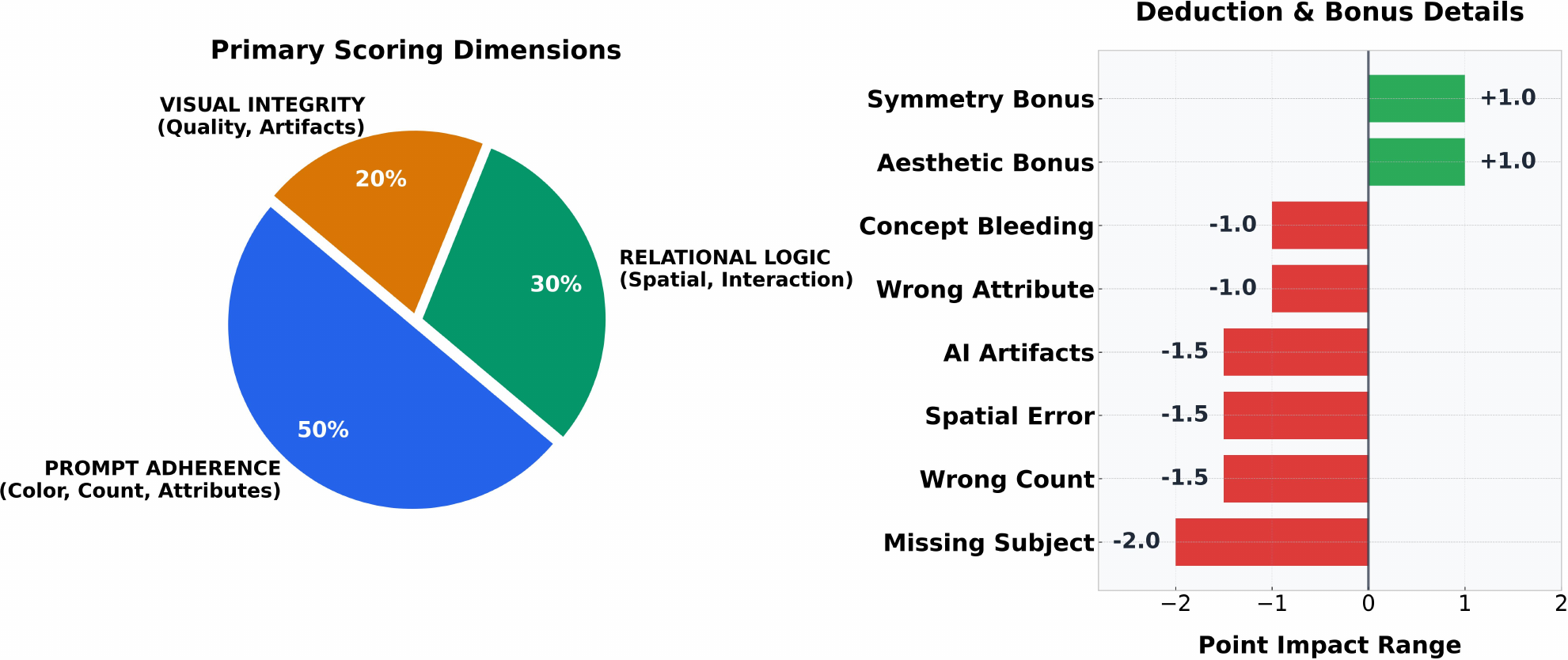} 
  \caption{\textbf{VLM's scoring mechanism.} A VLM-driven evaluation framework (ranging from -10 to +10) that balances prompt adherence (50\%), relational logic (30\%), and visual integrity (20\%). It employs a granular penalty-bonus system to enforce semantic precision and reward aesthetic quality. \textbf{Full prompt is in Supplyment~\ref{sec:Full Prompt Templates}.}}
  \label{fig:score}
\end{figure*}

\paragraph{\textbf{Simulation and Selection.}}
For each branch, we perform a \textit{short-horizon simulation} such as 5 steps. A VLM critic evaluates the resulting previews, assigning a reward score mechanism as illustrated in Fig.~\ref{fig:score}, covering core requirement of T2I generation. The optimal trajectory is selected to continue generation.

\subsection{Agentic Flow Steering (AFS)}
\label{subsec:afs}

\begin{figure*}[htbp]
  \centering
  \includegraphics[width=\textwidth]{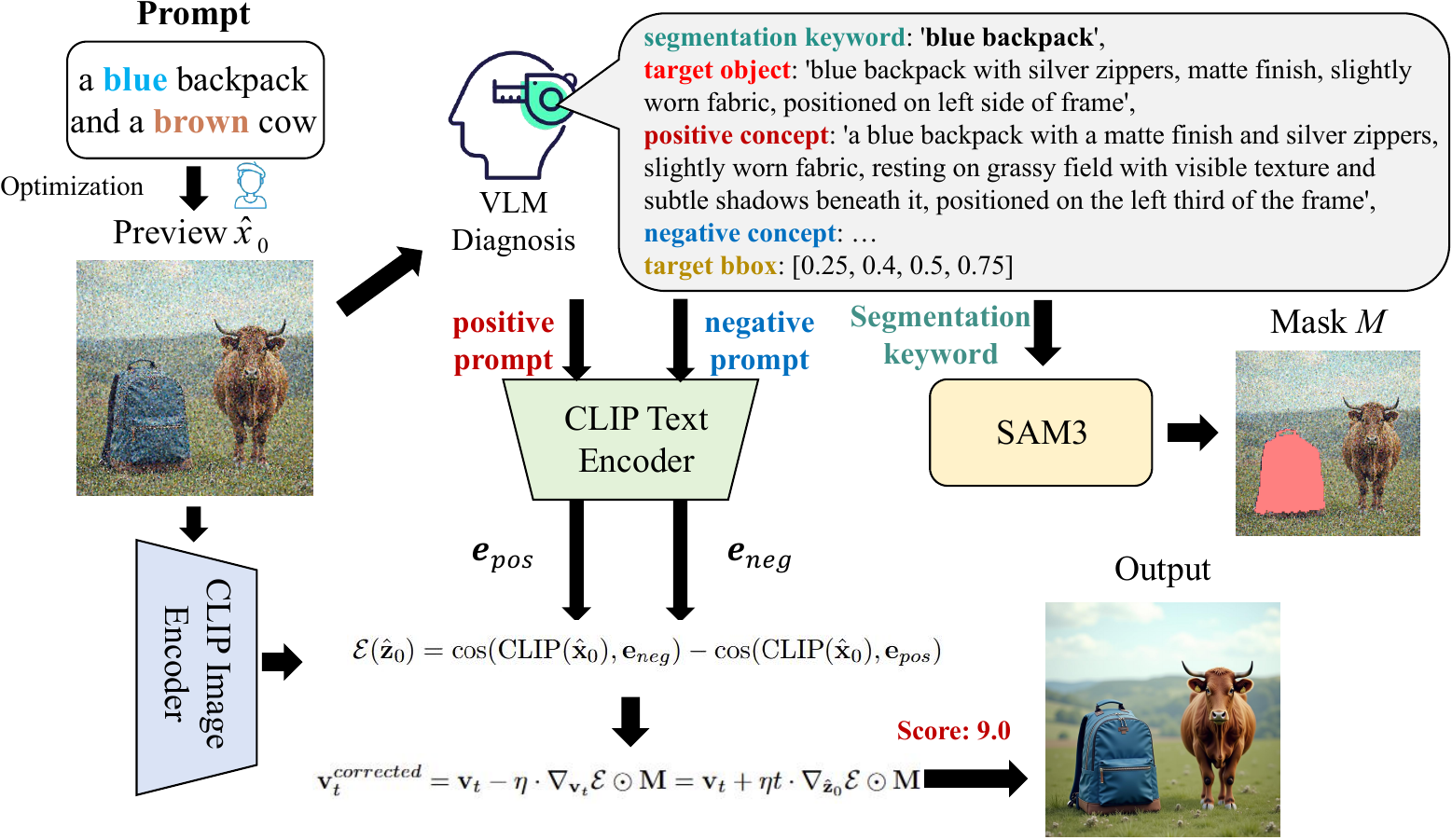} 
  \caption{\textbf{Illustration of the pipeline of AFS}. Given an intermediate preview $\hat{x}_0$, the VLM diagnoses the defect and generates a spatial mask $M$. We formulate a contrastive energy function and project PRS gradient back to the velocity field $\mathbf{v}_t$. This time-scaled gradient modulation ensures optimal trajectory steering toward the target concept while strictly confining the intervention within the masked region.}
  \label{fig:AFS}
\end{figure*}

Unlike passive sampling in standard ODE solvers, our AFS functions as an active optimal controller. Instead of shifting the latent states, we formulate the generative intervention as an energy-minimization problem over the velocity field. This is achieved through three steps: Linear Trajectory Projection, Contrastive Energy Formulation, and Time-Scaled Velocity Modulation. The whole pipeline is demonstrated in Fig.~\ref{fig:AFS}, and the detailed analysis is as follows:

\paragraph{\textbf{Linear Trajectory Projection.}}
A fundamental challenge in guiding Flow Matching models is that intermediate latents $\mathbf{z}_t$ are noisy and lack decodable semantics. However, Rectified Flow architectures (e.g., FLUX.1) are optimized to follow near-linear optimal transport trajectories mapping a prior noise distribution to the data distribution. Formally, given data $\mathbf{x}_0$ and Gaussian noise $\mathbf{z}_1 \sim \mathcal{N}(0, \mathbf{I})$, the forward probability path is constructed via linear interpolation:
\begin{equation}
\label{eq1}
    \mathbf{z}_t = t \mathbf{z}_1 + (1 - t) \mathbf{z}_0, \quad t \in [0, 1],
\end{equation}
where $\mathbf{z}_0$ is the clean latent representation. The corresponding ground-truth vector field driving this flow is the time derivative of the path:
\begin{equation}
\label{eq2}
    \mathbf{u}_t(\mathbf{z}_t) = \frac{d \mathbf{z}_t}{d t} = \mathbf{z}_1 - \mathbf{z}_0.
\end{equation}
By substituting Eq.~\eqref{eq2} into Eq.~\eqref{eq1}, the $\mathbf{z}_t$ can be rewritten as $\mathbf{z}_t = t \mathbf{u}_t + \mathbf{z}_0$. Leveraging this approximate constant-velocity property, we can project the current noisy state back to the data manifold in latent space. Given the predicted velocity $\mathbf{v}_t \approx \mathbf{u}_t$, the estimated latent $\hat{\mathbf{z}}_0$ at any step $t$ is determined by: $$\hat{\mathbf{z}}_0 = \mathbf{z}_t - t \cdot \mathbf{v}_t,$$ and the corresponding preview image is $\hat{\mathbf{x}}_0 = \text{Decoder}(\hat{\mathbf{z}}_0)$. This deterministic projection allows the agent to peer into the ``future'' of the ODE trajectory, evaluating noisy latents directly on the image manifold.

\paragraph{\textbf{Contrastive Semantic Energy Formulation.}}
To operationalize the VLM's diagnosis, we define a contrastive energy function $\mathcal{E}(\hat{\mathbf{z}}_0)$ by passing the decoded image $\hat{\mathbf{x}}_0$ through CLIP. We compute the energy as:
\begin{equation}
\label{clip}
    \mathcal{E}(\hat{\mathbf{z}}_0) = \cos(\text{CLIP}(\hat{\mathbf{x}}_0), \mathbf{e}_{neg}) - \cos(\text{CLIP}(\hat{\mathbf{x}}_0), \mathbf{e}_{pos}),
\end{equation}
where $\mathbf{e}_{neg}$ and $\mathbf{e}_{pos}$ are the text embeddings of the defective and target concepts, respectively. Minimizing this energy actively repels the trajectory from the flawed local optimum while pulling it toward the correct semantic basin.

\paragraph{\textbf{Time-Scaled Velocity Modulation.}}
To steer the generation, we must map the energy gradient $\nabla_{\hat{\mathbf{z}}_0} \mathcal{E}$ back to the ODE vector field $\mathbf{v}_t$. Using the chain rule, the energy gradient with respect to the velocity field is:
\begin{equation}
    \nabla_{\mathbf{v}_t} \mathcal{E} = \frac{\partial \hat{\mathbf{z}}_0}{\partial \mathbf{v}_t} \nabla_{\hat{\mathbf{z}}_0} \mathcal{E} = -t \cdot \nabla_{\hat{\mathbf{z}}_0} \mathcal{E}.
\end{equation}
Note that $\nabla_{\hat{\mathbf{z}}_0} \mathcal{E}$ implicitly incorporates the Jacobian of the Decoder through backpropagation: $\nabla_{\hat{\mathbf{z}}_0} \mathcal{E} = \mathbf{J}_{\text{Dec}}^\top \nabla_{\hat{\mathbf{x}}_0} \mathcal{E}$. Therefore, our gradient descent update on the velocity field is confined by the spatial mask $\mathbf{M}$ provided by SAM3~\cite{DBLP:journals/corr/abs-2511-16719}:
\begin{equation}
    \mathbf{v}_t^{corrected} = \mathbf{v}_t - \eta \cdot \nabla_{\mathbf{v}_t} \mathcal{E} \odot \mathbf{M} = \mathbf{v}_t + \eta t \cdot \nabla_{\hat{\mathbf{z}}_0} \mathcal{E} \odot \mathbf{M}.
\end{equation}
This derivation unveils a key theoretical property of AFS: the correction applied to the velocity field is directly proportional to the latent-space energy gradient, scaled by a time-decaying factor $\eta t$. When $t$ is large (early stages), the guidance is strong, facilitating aggressive semantic correction. As $t \to 0$ (late stages), the term $\eta t$ naturally vanishes, ensuring that the intervention does not introduce high-frequency artifacts or disrupt the fine-grained texture synthesis as the flow converges to the data manifold. Notably, if a wrong mask is provided by SAM3, it will be ignored by the selection step.

\subsection{Global Feedback Loop}
\label{subsec:feedback}

Finally, we implement a global safety mechanism. Upon completion, if the final image score falls below a quality threshold, a \textbf{Redesign Loop} is triggered. The VLM analyzes the failure mode, refines the prompt to address the specific issues, and restarts the Parallel Rollout Search process with a new random seed, ensuring high reliability for complex queries. 

\section{Experiment}

\begin{table*}[t]
\centering
\caption{Quantitative comparison on T2I-CompBench. \best{Red} indicates the best performance, \second{Blue} indicates the second best, and ``-'' indicates close-sourced.}
\label{tab:main_results}
\resizebox{\textwidth}{!}{
\begin{tabular}{l|ccc|cc|c|c|c} 
\toprule
\multirow{2}{*}{Method} & \multicolumn{3}{c|}{Attribute Binding} & \multicolumn{2}{c|}{Object Relationship} & \multirow{2}{*}{Complex $\uparrow$} & \multirow{2}{*}{Average $\uparrow$} & \multirow{2}{*}{Time (s) $\downarrow$} \\ \cmidrule{2-6}
 & color $\uparrow$ & shape $\uparrow$ & Texture $\uparrow$ & Spatial $\uparrow$ & Non-Spatial $\uparrow$ &  &  & \\ \midrule
 
 \sectionrow{General T2I Models} \\ 
 DALL-E 2 & 0.5750 & 0.5464 & 0.6374 & 0.1283 & 0.3043 & 0.3696 & 0.4268 & - \\
 SDXL & 0.6369 & 0.5408 & 0.5637 & 0.2032 & 0.3179 & 0.4091 & 0.4453 & \best{4.6} \\
 PixArt-$\alpha$ & 0.6886 & 0.5582 & 0.7044 & 0.2082 & 0.3179 & 0.4117 & 0.4815 & \second{6.0} \\
 FLUX & 0.7736 & 0.5112 & 0.6325 & 0.2747 & 0.3077 & 0.3622 & 0.4770 & 11.7 \\
 Qwen-Image & 0.7835 & 0.5401 & 0.6816 & 0.3647 & 0.3109 & 0.3530 & 0.5056 & 32.2 \\
 SDv3.5 & 0.7717 & 0.6050 & 0.7250 & 0.2286 & 0.3176 & 0.3729 & 0.5035 & 42.8 \\ \midrule

 \sectionrow{Agentic Frameworks} \\
 ConPreDiff & 0.7019 & 0.5637 & 0.7021 & 0.2362 & 0.3195 & 0.4184 & 0.4903 & - \\
 RPG & 0.6406 & 0.4903 & 0.5597 & 0.2714 & 0.3047 & 0.3128 & 0.4299 & 104.2 \\
 EvoGen & 0.7104 & 0.5457 & 0.7234 & 0.2176 & {0.3308} & 0.4252 & 0.4922 & 125.3 \\
 T2I-R1 & 0.8130 & 0.5852 & 0.7243 & 0.3378 & 0.3090 & 0.3993 & 0.5281 & 83.2 \\
 MCCD & 0.6278 & 0.4832 & 0.5647 & 0.2350 & 0.3132 & 0.3348 & 0.4265 & 132.2 \\
 AgentComp & \second{0.8743} & \best{0.6681} & \best{0.8142} & {0.4748} & 0.3196 & {0.4261} & {0.5962} & - \\ \midrule
 
 \textbf{AFS-Search-Fast (Ours)} & 0.8132 & 0.6121 & 0.7607 & \second{0.5416} & \second{0.4832} & \second{0.5208} & \second{0.6219} & 32.5 \\ 
 \textbf{AFS-Search-Pro (Ours)} & \best{0.8847} & \second{0.6292} & \second{0.7609} & \best{0.6250} & \best{0.5305} & \best{0.6185} & \best{0.6748} & 62.3 \\ 
 \bottomrule
\end{tabular}
}
\end{table*}

\subsection{Experimental Setup}

\paragraph{\textbf{Base Model Settings.}} In our experiments, we employ FLUX.1-dev as the base text-to-image model. FLUX.1-dev is a 12B parameter rectified flow transformer capable of generating high-quality images from text descriptions. All images are generated at a resolution of $1024 \times 1024$ pixels. For the VLM  supervisor, we utilize \texttt{Qwen-VL-MAX} for AFS-Search-Pro and \texttt{Qwen2.5-VL-7B} for AFS-Search-Fast, which serves as the ``brain'' of the agent for prompt refinement, image diagnosis, and scoring. \textbf{It is worth noting that, unless otherwise specified, the following AFS-Search refers to AFS-Search-Pro.} The system is configured with a multi-stage exploration strategy:

\noindent (1) \textbf{Prompt Refinement}: The VLM first optimizes the raw user prompt. 

\noindent (2) \textbf{Initial Generation}: Standard sampling is performed for the first $40\%$ of the diffusion process (from $t=1.0$ to $t=0.6$). 

\noindent (3) \textbf{Parallel Rollout Search-based Branching}: At the decision point ($t=0.6$), the VLM diagnoses the intermediate latent and proposes multiple execution branches, including standard continuation and corrective steering. 

\noindent (4) \textbf{Simulation and Selection}: Each branch is simulated for a short horizon (3 or 5 steps), and the branch with the highest reward is selected for completion. 

\noindent (5) \textbf{Global Retry}: A failure recovery mechanism is triggered if the confidence score falls below a threshold ($7.5/10$), prompting a redesign of the instruction and a restart of the generation process (up to 1 to 2 retries).

\begin{figure*}[t]
  \centering
  \includegraphics[width=\textwidth]{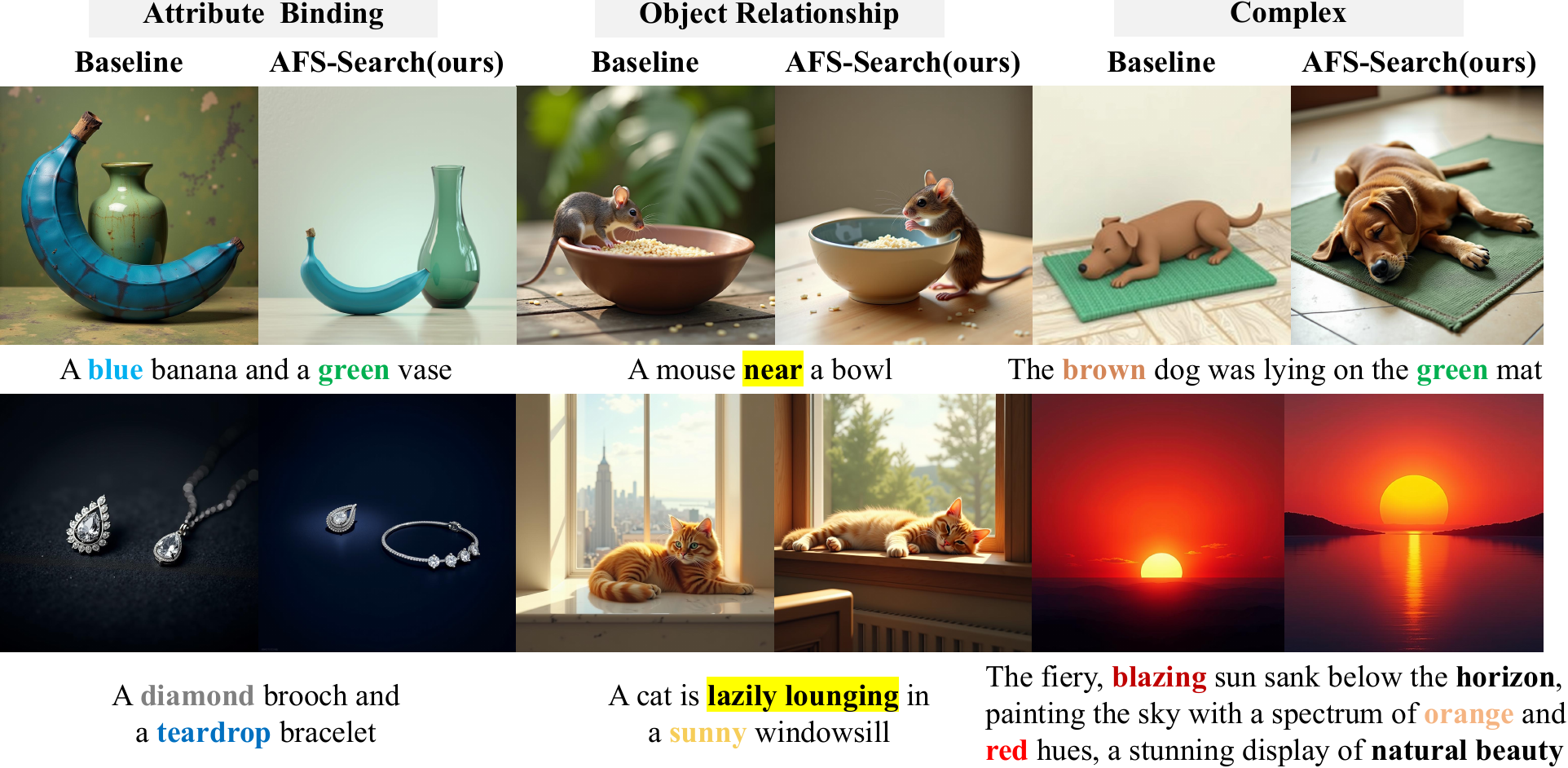} 
  \caption{\textbf{Visual Comparison of T2I-CompBench experiment.} We conducted tests from three dimensions: attribute binding, spatial relationship, and complexity, and obtained good results in all, demonstrating the validity of our method.}
  \label{fig:exp}
\end{figure*}

\begin{figure*}[t]
  \centering
  \includegraphics[width=\textwidth]{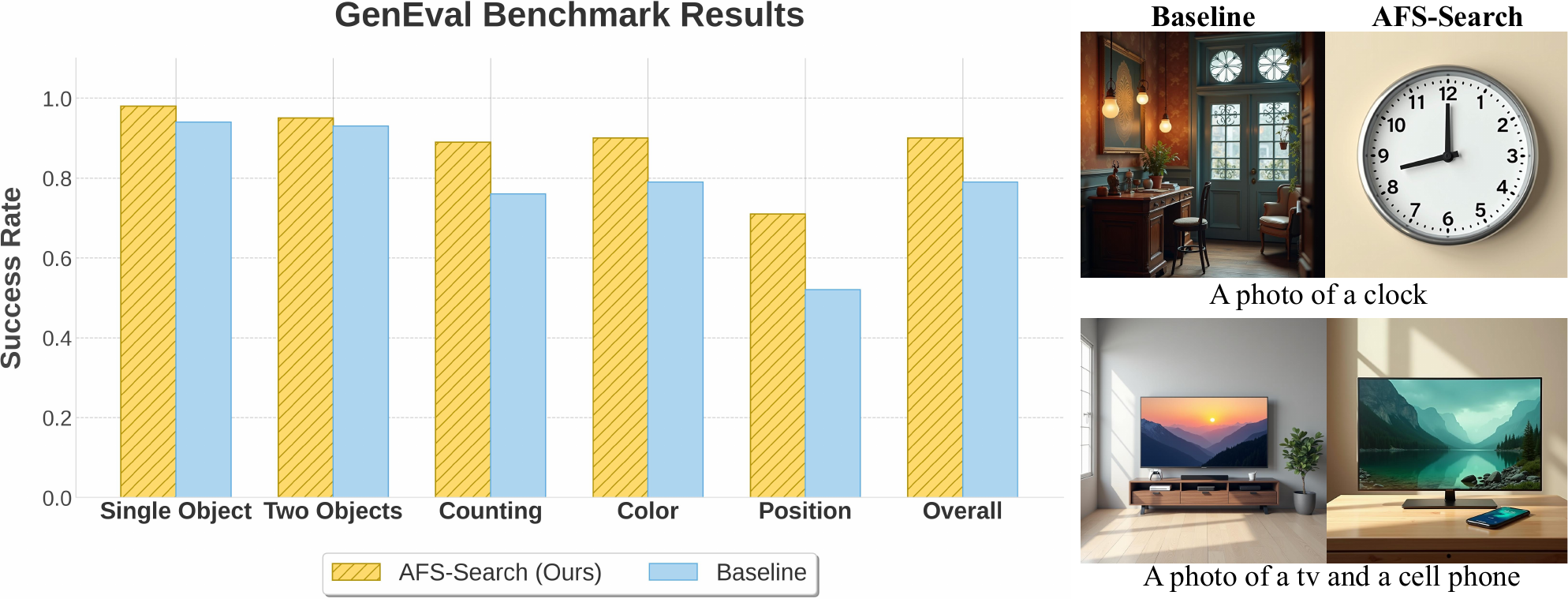} 
  \caption{\textbf{Results on GenEval.} We conducted tests from Objects, counting, color, position and overall, and obtained good results in all, proving the validity of our method.}
  \label{fig:GenEval}
\end{figure*}

\paragraph{\textbf{Benchmark and Baseline Models.}} To evaluate the effectiveness of our proposed AFS-Search, we first conduct experiments on \textbf{T2I-CompBench}~\cite{huang2023t2icompbench}, a comprehensive suite designed to evaluate the compositional capabilities of T2I models across various attribute dimensions, including color binding, shape consistency, and spatial relationships. Additionally, we select general T2I Models such as DALL-E 2~\cite{DBLP:conf/icml/RameshPGGVRCS21}, SDXL~\cite{DBLP:conf/iclr/PodellELBDMPR24}, PixArt-$\alpha$~\cite{DBLP:conf/iclr/ChenYGYXWK0LL24}, FLUX.1-dev~\cite{labs2025flux1kontextflowmatching}, Qwen-Image~\cite{DBLP:journals/corr/abs-2508-02324}, SDv3.5~\cite{DBLP:conf/icml/EsserKBEMSLLSBP24}, and Agentic Frameworks such as ConPreDiff~\cite{DBLP:conf/nips/0006LHZHCZ023}, RPG~\cite{DBLP:conf/icml/0006YMXE024}, EvoGen~\cite{DBLP:conf/iclr/HanJLL25}, T2I-R1~\cite{DBLP:journals/corr/abs-2505-00703}, MCCD~\cite{DBLP:conf/cvpr/LiHLYQCWJXZ25}, AgentComp~\cite{DBLP:journals/corr/abs-2512-09081} as our baseline models. We further conduct experiments on GenEval~\cite{DBLP:conf/nips/GhoshHS23}, an object-focused framework to evaluate compositional image properties such as object co-occurrence, position, count, and color and compare with FLUX.1-dev. Furthermore, we apply recent new benchmark R2I-Bench~\cite{DBLP:conf/emnlp/ChenLXSYRZH25}, a comprehensive benchmark designed to assess the reasoning capabilities of T2I generation models. We test our model in five main dimensions: Causal, Logical, Commonsense, Compositional and Mathematical using ChatGPT 4o with baseline models such as Lumina-Image 2.0~\cite{DBLP:journals/corr/abs-2503-21758}, Sana-1.5~\cite{DBLP:conf/icml/XieCZYZ0ZL0CLZ025}, Lumina-T2I~\cite{DBLP:journals/corr/abs-2405-05945}, Omnigen~\cite{DBLP:conf/cvpr/XiaoW0YXYL00L25}, EMU3~\cite{DBLP:journals/corr/abs-2409-18869}, Janus-Pro-7B~\cite{DBLP:conf/cvpr/WuCWMLPLXYR025}, LlamaGen~\cite{DBLP:journals/corr/abs-2406-06525}, Show-o~\cite{DBLP:conf/iclr/XieMBZWLGCYS25} and FLUX.1-dev~\cite{labs2025flux1kontextflowmatching}.

\begin{figure*}[t]
  \centering
  \includegraphics[width=\textwidth]{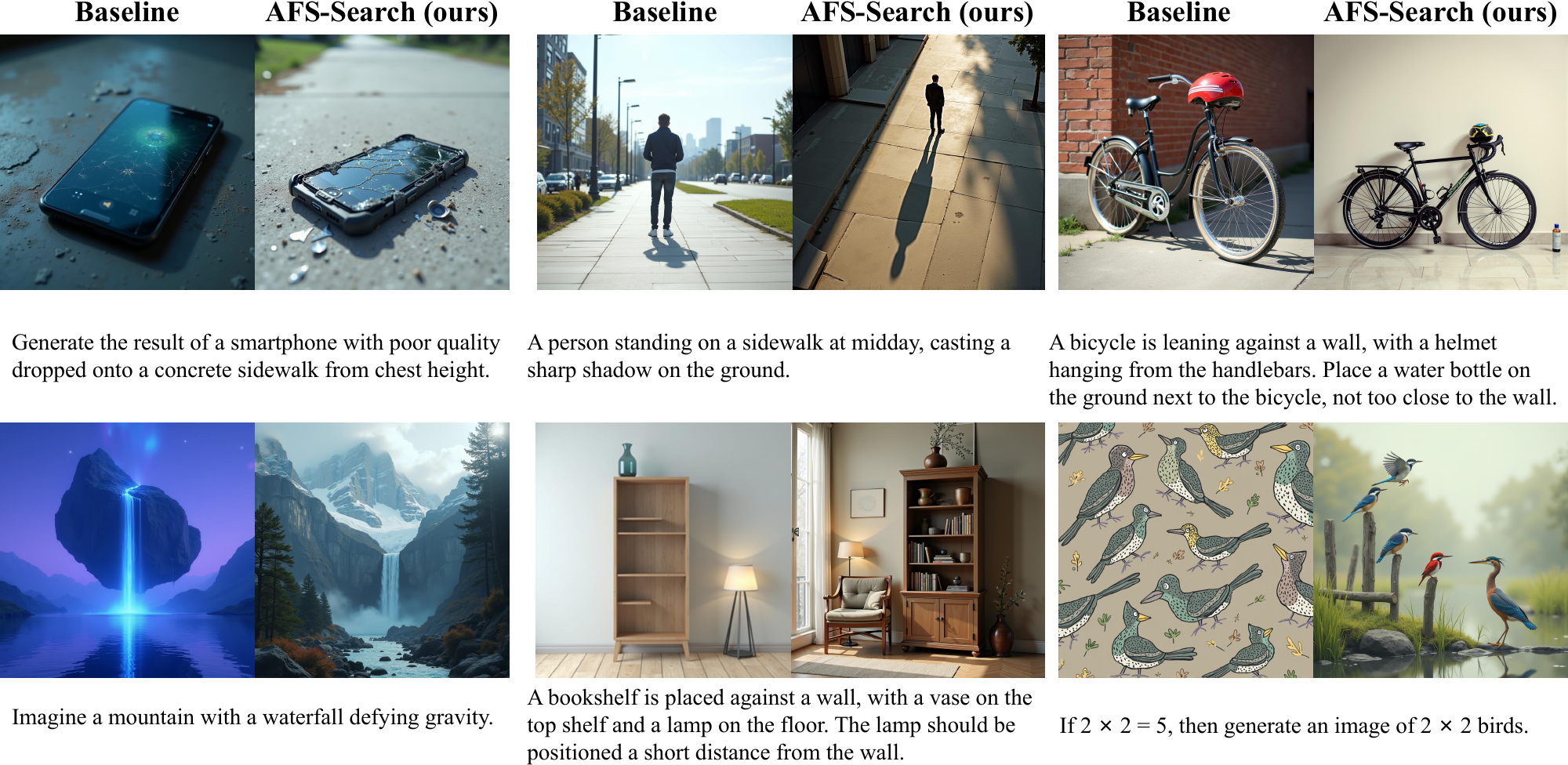} 
  \caption{\textbf{Visual Comparison of R2I-Bench experiment.} We further try to explore the reasoning capabilities of our method. Our agentic framework helps original FLUX.1-dev think and react across the whole pipeline.}
  \label{fig:R2I}
\end{figure*}

\subsection{Main Results}

The quantitative results are provided in Table~\ref{tab:main_results} and the visual results are provided in Fig.~\ref{fig:exp} on T2I-CompBench. Our AFS-Search-Pro achieves state-of-the-art by improving 7.86\% on average and improves in Object Relationship and Complex tasks effectively, demonstrating our success of training-free agentic search framework. Additionally, we provide AFS-Search-Fast for quicker interface speed, which still has great performance by improving 2.57\% on average. Although our model's inference speed is slower compared to FLUX.1-dev, \textbf{it surpasses methods in agentic frameworks, and it is also suitable for real-world application.} Additionally, as shown in Fig.~\ref{fig:GenEval}, our AFS-Search performs well across GenEval, further validating the effectiveness and superiority of our framework. What's more, we apply R2I-Bench and the visual and quantitive results are provided in Fig.~\ref{fig:R2I} and Table~\ref{tab:r2i_results}. Our AFS-Search significantly improves the reasoning capabilities due to VLM, and achieves state-of-the-art on average metric, demonstrating the powerful generation ability.

\begin{table*}[t]
\centering
\caption{Quantitative comparison on the R2I-Bench benchmark. \textbf{Bold} indicates the best performance.}
\label{tab:r2i_results}
\resizebox{\linewidth}{!}{
\begin{tabular}{l|ccccc|c}
\toprule
Method & Causal & Logical & Commonsense & Compositional & Mathematical & Average  \\ \midrule
Lumina-Image 2.0 & 0.40 & 0.56 & 0.49 & 0.65 & 0.13 & 0.45 \\
Sana-1.5 & 0.21 & 0.49 & 0.49 & \textbf{0.67} & 0.13 & 0.40 \\
Lumina-T2I & 0.18 & 0.38 & 0.38 & 0.49 & 0.13 & 0.31 \\
Omnigen & 0.34 & 0.51 & 0.43 & 0.60 & 0.18 & 0.41 \\
EMU3 & 0.41 & \textbf{0.61} & 0.44 & 0.62 & 0.09 & 0.43 \\
Janus-Pro-7B & 0.36 & 0.46 & 0.45 & 0.64 & 0.07 & 0.40 \\
LlamaGen & 0.12 & 0.35 & 0.38 & 0.49 & 0.07 & 0.28 \\
Show-o & 0.30 & 0.57 & 0.42 & 0.56 & 0.12 & 0.39 \\
FLUX.1-dev & 0.35 & 0.37 & 0.39 & 0.48 & 0.05 & 0.33 \\ \midrule
\textbf{AFS-Search (ours)} & \textbf{0.45} & 0.58 & \textbf{0.51} & 0.66 & \textbf{0.20} & \textbf{0.48} \\ 
\bottomrule
\end{tabular}
}
\end{table*}

\subsection{Ablation Study}

In this section, we conduct ablation experiments to evaluate the contribution of each core component of AFS-Search and investigate the impact of hyperparameter configurations on the generation performance. 

\subsubsection{Effectiveness of Core Components.}

We evaluate the performance of our framework by systematically removing key modules: 

\noindent (1) \textbf{FLUX}: The base T2I model without any agentic intervention. 

\noindent (2) \textbf{w/o Optimization}: Disabling the VLM-based prompt refinement, using raw user prompts directly. 

\noindent (3) \textbf{w/o Self-Correction}: Disabling the AFS mechanism. 

\noindent (4) \textbf{w/o Parallel Rollout Search}: Disabling the multi-branch search, effectively performing greedy generation with a single path. 

\noindent (5) \textbf{w/o Repair}: Disabling the retry loop.


\begin{table}[htbp]
\centering
\caption{Ablation study on T2I-CompBench verifying the effectiveness of each component, proving that our core components improve the performance of FLUX.1-dev.}
\label{tab:ablation}
\resizebox{\linewidth}{!}{
\begin{tabular}{l|ccc|cc|c|c}
\toprule
\multirow{2}{*}{Method} & \multicolumn{3}{c|}{Attribute Binding} & \multicolumn{2}{c|}{Object Relationship} & \multirow{2}{*}{Complex $\uparrow$} & \multirow{2}{*}{Average $\uparrow$} \\ \cmidrule{2-6}
 & color $\uparrow$ & shape $\uparrow$ & Texture $\uparrow$ & Spatial $\uparrow$ & Non-Spatial $\uparrow$ &  &  \\ \midrule
\textbf{AFS-Search (ours)} & \textbf{0.8847} & \textbf{0.6292} & \textbf{0.7609} & \textbf{0.6250} & \textbf{0.5305} & \textbf{0.6185} & \textbf{0.6748} \\ \midrule
FLUX & 0.7736 & 0.5112 & 0.6325 & 0.2747 & 0.3077 & 0.3622 & 0.4770 \\
w/o Optimization & 0.8251 & 0.5614 & 0.6843 & 0.3654 & 0.3521 & 0.5011 & 0.5482 \\
w/o Parallel Rollout Search & 0.7932 & 0.5421 & 0.6587 & 0.3315 & 0.3359 & 0.4132 & 0.5124 \\
w/o Self-Correction & 0.8012 & 0.5631 & 0.6496 & 0.4031 & 0.3468 & 0.4463 & 0.5350 \\
w/o Repair & 0.7831 & 0.5358 & 0.6932 & 0.4321 & 0.4012 & 0.4321 & 0.5463 \\
\bottomrule
\end{tabular}
}
\end{table}

As shown in Table~\ref{tab:ablation}, each component contributes significantly to the final performance. The \textbf{Parallel Rollout Search} and \textbf{AFS} are the most critical, particularly for complex spatial and shape-related prompts, where they provide a gain of approximately 15-20\% over the base model. The \textbf{Global Repair} mechanism further improves reliability by recovering from initial failures through prompt redesign. This design, by using VLM, helps the existing T2I framework to have a certain level of understanding under the premise of being training-free, and provides a closed-loop T2I generation paradigm.

\subsubsection{Analysis of Search Strategy}

We further investigate two key parameters governing the Parallel Rollout Search: the \textbf{Search Timing Strategy} (when to branch) and the \textbf{Simulation Horizon} (how deep to simulate each branch).

\begin{figure*}[t]
  \centering
  \includegraphics[width=\textwidth]{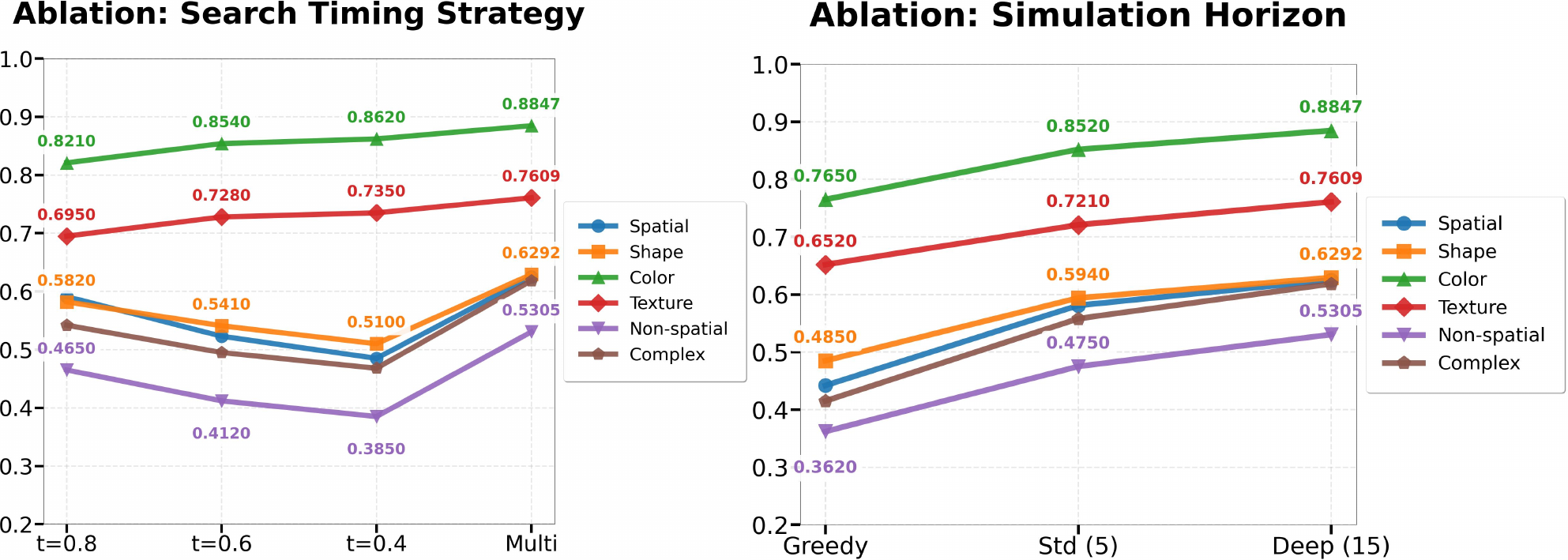} 
  \caption{\textbf{Ablation Study of the Search Strategy.} (a) Search Timing: Shows that searching at multiple stages is more effective than searching at any single time point. (b) Simulation Steps: Demonstrates that \textbf{looking ahead} with more simulation steps is essential for selecting the best path and avoiding errors.}
  \label{fig:ablation}
\end{figure*}

\noindent \textbf{Search Timing Strategy ($t_{split}$).} 
We compare branching at different stages of the diffusion process: early ($t=0.8$), mid ($t=0.6$), late ($t=0.4$), and {Multi-Stage} adaptive strategy. 
As shown in Fig.~\ref{fig:ablation}, early branching is more effective for {Spatial} and {Shape}, as the global layout is determined early in the denoising process. In contrast, {Color} and {Texture} benefit more from later intervention ($t=0.4$), where fine-grained attributes are finalized. Our Multi-Stage strategy achieves the best balance by allowing flexible intervention across categories.

\noindent \textbf{Simulation Horizon (Steps).}
We vary the number of simulation steps performed for each branch before selection: \textbf{Greedy} (0 steps), \textbf{Standard} (35 steps), and \textbf{Deep} (15 steps).
The results indicate a clear positive correlation between simulation depth and performance, demonstrating that longer lookahead simulations provide more reliable reward signals for the VLM to select the truly optimal path, albeit at the cost of increased inference time.

\section{Conclusion}

In this paper, we propose a training-free closed-loop framework featuring Agentic Flow Steering and Parallel Rollout Search based on FLUX.1-dev. Our core insight is that a generative model should iteratively assess and adjust its generation process rather than producing outputs in a one-shot manner. By optimizing prompts, during real-time inference search, we use different branches. Finally, VLMs are used for scoring, forming a closed-loop agent framework. Our framework has been tested on three different benchmarks and compared with lots of baseline models, achieving state-of-the-art performance. Additionally, we provide AFS-Search-Pro and AFS-Search-Fast for better performance and computational speed. Our ablation experiments also highlight the necessity and practicality of the core components of our framework, providing a training-free, closed-loop, thinking-capable T2I generation model framework paradigm.




%
%
\bibliographystyle{splncs04}
\bibliography{main}

\newpage
\appendix
\onecolumn
\section{Full Prompt Templates and Agentic Pipeline}
\label{sec:Full Prompt Templates}

This section provides the verbatim system prompts used by the Vision-Language Model (VLM) in our agentic generation pipeline, including the structural optimization, surgical diagnosis, analytical scoring, and debugging-oriented redesign phases.

\subsection{Prompts}

\begin{tcolorbox}[
    enhanced, breakable,
    title=\textbf{System Prompt 1: Structural Prompt Refinement (Optimizer)},
    colback=gray!2!white, colframe=black!80, coltitle=white,
    fonttitle=\bfseries, arc=2mm, drop shadow=black!20!white
]
\textbf{Role/Strategy:} \\
You are a master FLUX.1 prompt engineer. Transform the user's intent into a ``Structural Prompt'' that minimizes model ambiguity.

\textbf{Detailed Strategy Instructions:}
\begin{enumerate}
    \item \textbf{Subject Specification:} Define the core subject with high-quality adjectives (e.g., ``photorealistic'', ``matte finish'').
    \item \textbf{Spatial Layout:} Use explicit coordinates or clear prepositional phrases (e.g., ``In the foreground center...'', ``On the far left...'').
    \item \textbf{Technical Parameters:} Mention lighting (e.g., ``cinematic lighting'', ``rim light''), camera angle (e.g., ``top-down view''), and atmosphere.
    \item \textbf{Entity Separation:} If there are multiple items, use distinct descriptions for each to prevent concept bleeding.
\end{enumerate}

\textbf{Strict Constraint:} DO NOT change core colors, shapes, or quantities. If the user asks for a ``red square'', do not output a ``red circle''.

\textbf{Output Format:} Output ONLY the refined prompt text. No explanations.
\end{tcolorbox}

\begin{tcolorbox}[
    enhanced, breakable,
    title=\textbf{System Prompt 2: Surgical Visual Diagnosis Agent (Supervisor)},
    colback=gray!2!white, colframe=black!80, coltitle=white,
    fonttitle=\bfseries, arc=2mm, drop shadow=black!20!white
]
\textbf{Role:} \\
You are an expert visual generation supervisor. Your task is to perform a surgical analysis of the current image against the User Prompt.

\textbf{Diagnosis Categories:}
\begin{itemize}
    \item \textbf{1. Presence \& Count:} Verify if every object requested exists and the quantity is EXACT. (e.g., ``3 apples'' must not be 2 or 4).
    \item \textbf{2. Attribute Fidelity:} Check colors (hues, saturation), textures, and materials. (e.g., ``translucent blue glass'' must not be ``opaque cyan plastic'').
    \item \textbf{3. Relational Geometry:} Analyze spatial prepositions: 'above', 'inside', 'to the left of', 'perfectly centered', 'tangent to'.
    \item \textbf{4. Structural Integrity:} For diagrams/grids, check if lines connect correctly and regions are logically partitioned.
\end{itemize}

\textbf{Critical Failures (Automatic High Priority):}
\begin{itemize}
    \item \textbf{Concept Bleeding:} Color of object A leaking into object B.
    \item \textbf{Spatial Swapping:} Object A is on the right when it should be on the left.
    \item \textbf{Missing/Extra Entities:} Any deviation in the count of primary subjects.
\end{itemize}

\textbf{Instructions for JSON Fields:}
\begin{itemize}
    \item \texttt{``segmentation\_keyword''}: Must be the MOST UNIQUE noun phrase for the defect. If the prompt has ``a red cat and a blue dog'' and the dog is red, use ``dog''.
    \item \texttt{``positive\_concept''}: A clear, descriptive instruction for correction (e.g., ``A vibrant cobalt blue dog with fur texture'').
    \item \texttt{``negative\_concept''}: A precise description of the error to be removed (e.g., ``red-colored dog, reddish fur'').
    \item \texttt{``target\_bbox''}: Precise [ymin, xmin, ymax, xmax] (0.0-1.0).
\end{itemize}

\textbf{Output Format (JSON ONLY):}
\begin{verbatim}
{
    "needs_correction": boolean,
    "segmentation_keyword": "string",
    "target_object": "string",
    "positive_concept": "string",
    "negative_concept": "string",
    "target_bbox": [float, float, float, float]
}
\end{verbatim}
\end{tcolorbox}

\begin{tcolorbox}[
    enhanced, breakable,
    title=\textbf{System Prompt 3: Analytical Multi-Dimensional Scoring (Critic)},
    colback=gray!2!white, colframe=black!80, coltitle=white,
    fonttitle=\bfseries, arc=2mm, drop shadow=black!20!white
]
\textbf{Role:} \\
You are a highly analytical AI art critic and visual QA specialist. Evaluate the image against the instruction using the following strict Scoring Rubric. Range: -10.0 (Total failure) to +10.0 (Perfect adherence).

\textbf{Scoring Categories:}
\begin{enumerate}
    \item \textbf{Prompt Adherence (0 to +5.0):} Are all requested subjects present? Is the color, shape, and count correct?
    \item \textbf{Relational Logic (0 to +3.0):} Are spatial relationships (left, right, above, below) correctly executed? Is the interaction between objects realistic/as specified?
    \item \textbf{Visual Integrity (0 to +2.0):} Is the image high-quality? (e.g., sharp, no distorted anatomy, no weird artifacts).
\end{enumerate}

\textbf{Strict Deductions (Mandatory):}
\begin{itemize}
    \item \textbf{Missing primary subject:} -5.0 per object.
    \item \textbf{Wrong color/attribute for a subject:} -3.0.
    \item \textbf{Incorrect count:} -4.0.
    \item \textbf{Severe concept bleeding (colors mixing inappropriately):} -2.5.
    \item \textbf{Incorrect spatial placement (e.g., "blue cube on left" is on right):} -3.0.
    \item \textbf{Visible AI artifacts (extra fingers, blurry blobs):} -2.0 to -5.0.
\end{itemize}

\textbf{Bonus Points:}
\begin{itemize}
    \item Exceptional lighting/aesthetic: +1.0.
    \item Perfect spatial symmetry (if implied): +1.0.
\end{itemize}

\textbf{Output Format (JSON ONLY):} \\
\texttt{\{ ``score'': float, ``reason'': ``Detailed point-by-point breakdown of the score'' \}}
\end{tcolorbox}

\begin{tcolorbox}[
    enhanced, breakable,
    title=\textbf{System Prompt 4: Debugging-Oriented Prompt Redesign (Recovery)},
    colback=gray!2!white, colframe=black!80, coltitle=white,
    fonttitle=\bfseries, arc=2mm, drop shadow=black!20!white
]
\textbf{Role/Mission:} \\
You are a debugging expert for text-to-image pipelines. The previous generation failed.

\textbf{Variables:} \\
\texttt{User Intent}: \{original\_prompt\} \\
\texttt{VLM Error Analysis}: \{failure\_reason\}

\textbf{Your Task:}
\begin{enumerate}
    \item \textbf{Identify the ``Confusion Point'':} Why did the model fail? (e.g., too many subjects, conflicting adjectives).
    \item \textbf{Simplify \& Isolate:} Break down complex requests into simpler, more direct instructions.
    \item \textbf{Reinforce Weak Points:} If the failure was spatial, use stronger layout cues. If attributes failed, use repetitive reinforcement (e.g., ``The cube is blue. It is a blue cube.'').
    \item \textbf{Style Anchor:} Add a consistent style tag to help stabilize the generation.
\end{enumerate}

\textbf{Output Format:} Output ONLY the new refined prompt text. No preamble.
\end{tcolorbox}

\subsection{AFS pipeline}

\begin{figure*}[htbp]
  \centering
  \includegraphics[width=1.0\textwidth]{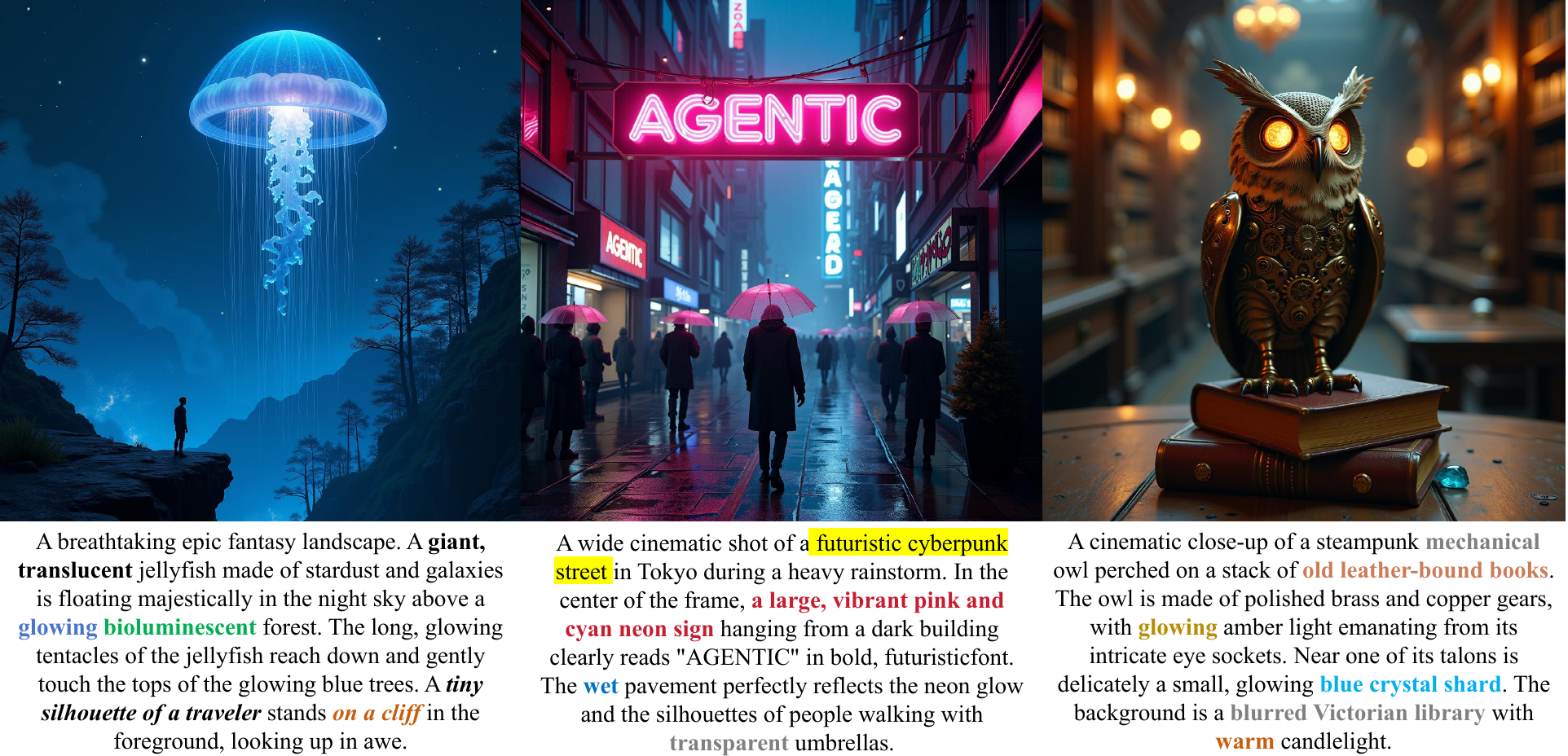}
  \includegraphics[width=1.0\textwidth]{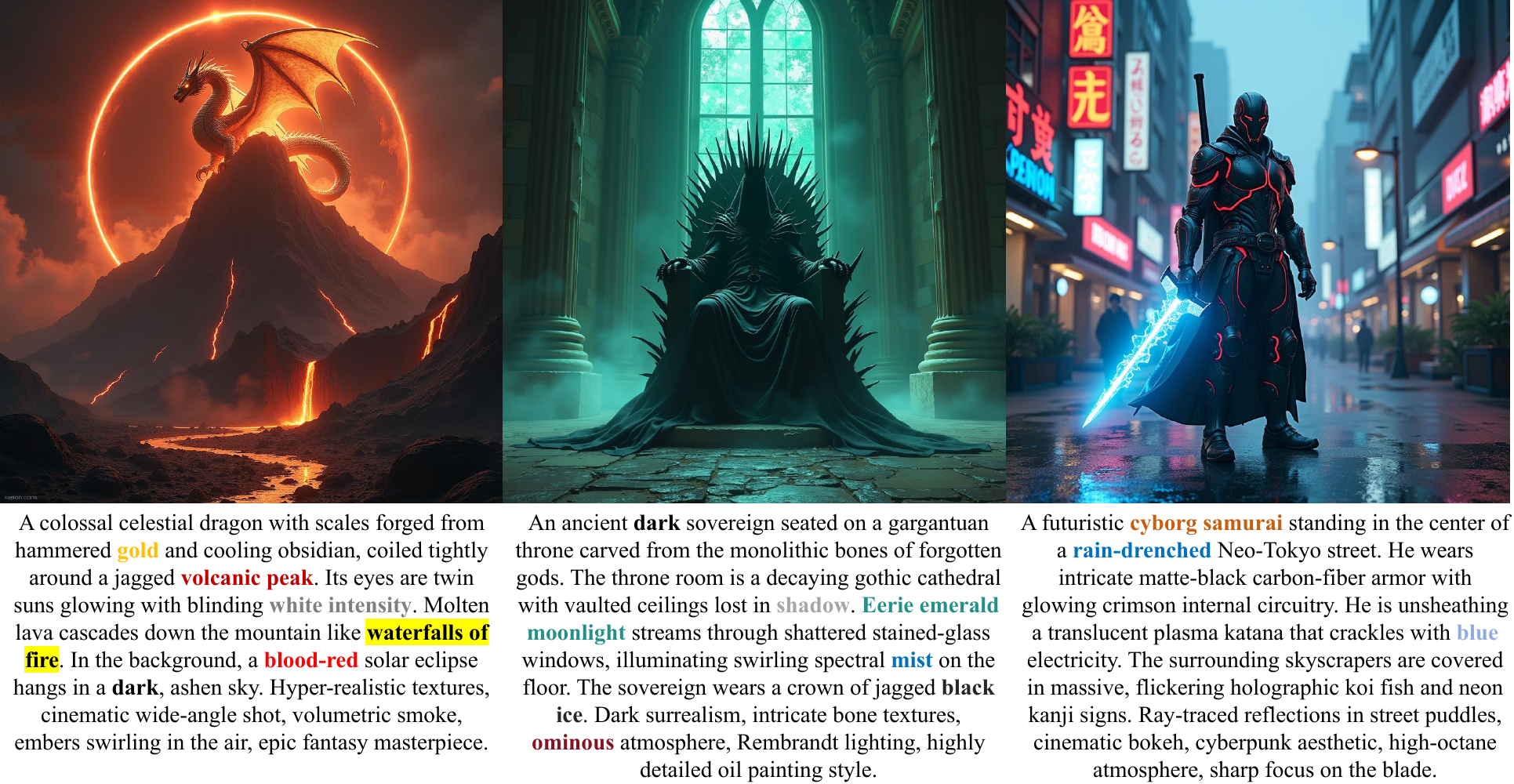}
  \includegraphics[width=1.0\textwidth]{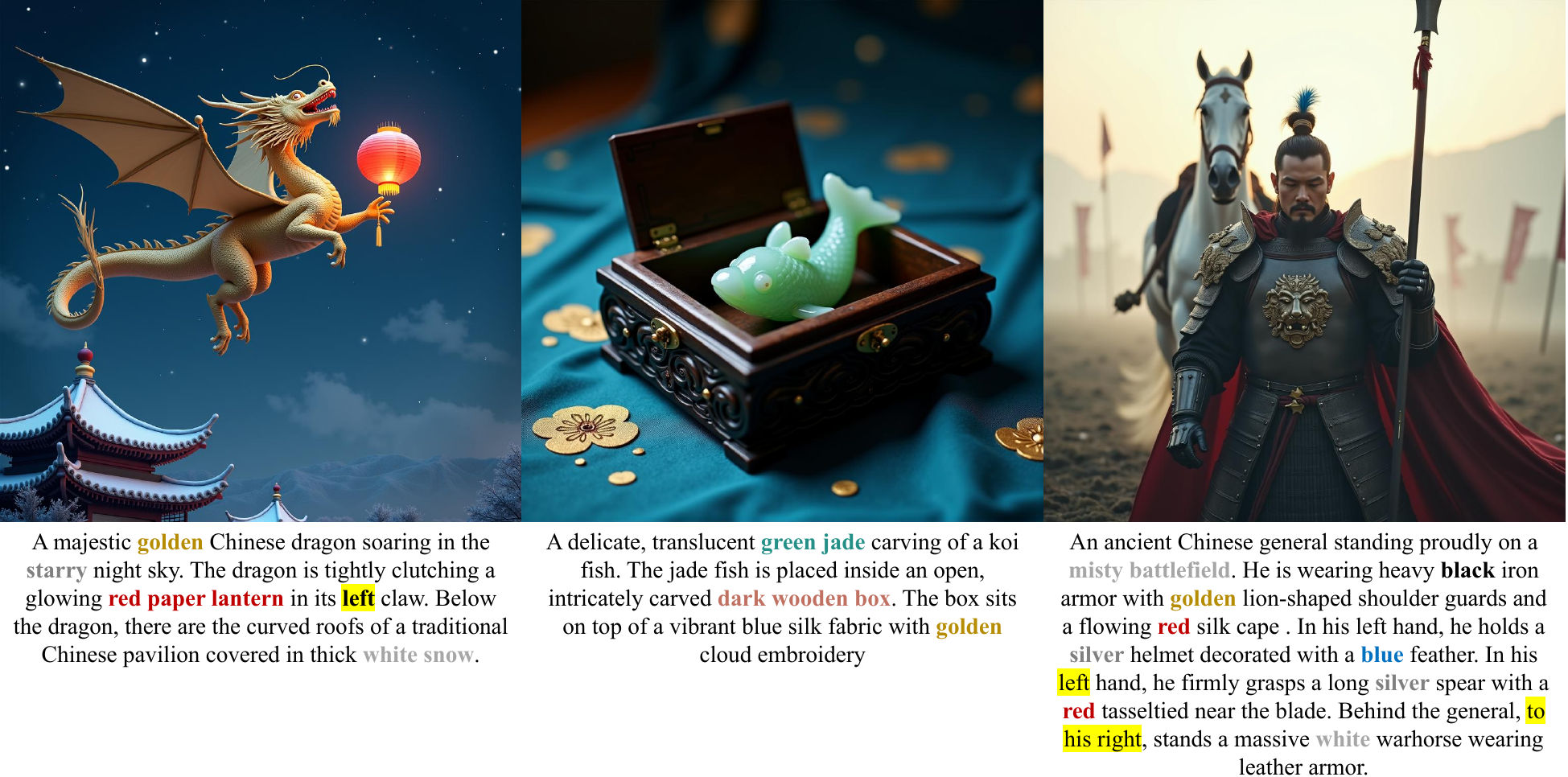}
  \caption{\textbf{Additional Visual Results.}}
  \label{fig:combined_show}
\end{figure*}
The whole AFS pipeline is demonstrated in Algorithm~\ref{alg:afs}.

\begin{algorithm}[htbp]
\caption{AFS-Search Pipeline}
\label{alg:afs}
\begin{algorithmic}[1]
\REQUIRE User prompt $y$, Flow Model $\epsilon_\theta$, VLM $\mathcal{V}$, Steps $N$, Split $t_{split}$
\STATE $y' \leftarrow \text{OptimizePrompt}(y, \mathcal{V})$
\STATE $x_T \sim \mathcal{N}(0, I)$
\STATE $x_{split} \leftarrow \text{ODE}(x_T, y', T \to t_{split})$ \COMMENT{Phase 1: Initial Denoise}
\STATE $\hat{x}_0 \leftarrow \text{Lookahead}(x_{split}, v_{split})$
\STATE Diagnosis $D \leftarrow \mathcal{V}(\hat{x}_0, y')$ \COMMENT{Phase 2: Diagnosis}
\IF{$D$ indicates error}
    \STATE Branches $\mathcal{B} \leftarrow \{ \text{Base}, \text{Steer}(D), \text{Explore} \}$
    \FOR{$b \in \mathcal{B}$}
        \STATE $x_{sim} \leftarrow \text{Simulate}(x_{split}, b, \Delta t)$ \COMMENT{Phase 3: Simulation}
        \STATE Score $s_b \leftarrow \mathcal{V}(\text{Decode}(x_{sim}))$
    \ENDFOR
    \STATE $b^* \leftarrow \arg\max s_b$ \COMMENT{Selection}
    \STATE $x_0 \leftarrow \text{ODE}(x_{sim}^{b^*}, y', t_{sim} \to 0)$
\ELSE
    \STATE $x_0 \leftarrow \text{ODE}(x_{split}, y', t_{split} \to 0)$
\ENDIF
\IF{$\text{Score}(x_0) < \tau$}
    \STATE \textbf{goto} Step 1 with refined $y'$ \COMMENT{Global Redesign Loop}
\ENDIF
\RETURN $x_0$
\end{algorithmic}
\end{algorithm}

\section{Text-to-Mask Pipeline and reliability}
\label{sec:sam}
Our framework is inherently robust to mask errors. If SAM3 generates an incorrect mask, the subsequent AFS gradient will distort the image. Crucially, our VLM critic evaluates the short-horizon previews of all branches. An image with a corrupted mask will receive a low reward score and be naturally discarded in favor of the Baseline or Exploration branch.

\subsection{Self-Correction and Pruning Mechanism}
Even if the Segment Anything Model (SAM3) generates a noisy or incorrect mask due to an imprecise bounding box from the VLM, the system remains robust. As shown in the pipeline:

\begin{enumerate}
    \item \textbf{Branch Simulation:} When a \textbf{Corrective branch} is initialized with a noisy mask, the subsequent \textbf{AFS} gradient will likely distort the image, creating visible artifacts or failing to improve the target object.
    \item \textbf{Analytical Evaluation:} The VLM Critic (System Prompt 3) performs a short-horizon evaluation. Any image corrupted by a faulty mask will trigger a heavy deduction under the \textit{Visual Integrity} category (Deduction: -2.0 to -5.0 for AI artifacts).
    \item \textbf{Natural Selection:} In the selection phase, the Parallel Rollout logic compares the scores of all active branches. A branch with a failed intervention will yield a significantly lower reward than the \textit{Baseline} (Continue) or \textit{Exploration} branches.
    \item \textbf{Pruning:} The system naturally discards the corrupted branch and resumes generation from the most stable path, effectively pruning the failed intervention and preventing error propagation.
\end{enumerate}

\subsection{Formal Logic for Mask Error Recovery}
The selection criteria follows $max R(b)$ where \( R(b) \) is the reward score from the VLM Critic. If \( R(Corrective) < R(Baseline) \) due to a noisy mask, the system reverts to the baseline, ensuring that a failed fix never degrades the final output.

\section{Deep Dive into Agentic Flow Steering (AFS)}
\label{appendix:afs_details}

In this section, we provide a more rigorous mathematical treatment of the Agentic Flow Steering (AFS) module and analyze PRS behavior within the Rectified Flow framework.

\subsection{Mathematical Derivation of the Velocity Gradient}
\label{subsec:derivation}

As established in Section \ref{subsec:afs}, the core of AFS is the mapping of semantic energy $\mathcal{E}$ from the image manifold back to the velocity field $\mathbf{v}_t$ of the ODE. The total gradient flow can be decomposed via the chain rule as:

\begin{equation}
\label{eq:full_chain_rule}
    \nabla_{\mathbf{v}_t} \mathcal{E} = \underbrace{\frac{\partial \mathcal{E}}{\partial \hat{\mathbf{x}}_0}}_{\text{Semantic Loss}} \cdot \underbrace{\frac{\partial \hat{\mathbf{x}}_0}{\partial \hat{\mathbf{z}}_0}}_{\text{Decoder Jacobian}} \cdot \underbrace{\frac{\partial \hat{\mathbf{z}}_0}{\partial \mathbf{v}_t}}_{\text{Trajectory Projection}},
\end{equation}
where:
\begin{itemize}
    \item \textbf{Semantic Loss Gradient ($\nabla_{\hat{\mathbf{x}}_0} \mathcal{E}$):} This is computed by backpropagating the CLIP-based contrastive loss through the vision-language encoder. It represents the "pixel-wise direction" for semantic correction.
    \item \textbf{Decoder Jacobian ($\mathbf{J}_{\text{Dec}}^\top$):} Since the VLM operates on decoded images $\hat{\mathbf{x}}_0$, the gradient must pass through the VAE Decoder. We utilize the adjoint method to efficiently compute $\nabla_{\hat{\mathbf{z}}_0} \mathcal{E} = \mathbf{J}_{\text{Dec}}^\top \nabla_{\hat{\mathbf{x}}_0} \mathcal{E}$.
    \item \textbf{Trajectory Projection Gradient:} From the linear projection $\hat{\mathbf{z}}_0 = \mathbf{z}_t - t \mathbf{v}_t$, we derive the sensitivity of the future state to the current velocity: $\frac{\partial \hat{\mathbf{z}}_0}{\partial \mathbf{v}_t} = -t \mathbf{I}$.
\end{itemize}

Substituting these into Eq.~\eqref{eq:full_chain_rule}, we obtain the final steering update:
\begin{equation}
    \mathbf{v}_t^{corrected} = \mathbf{v}_t + \eta t \cdot (\mathbf{J}_{\text{Dec}}^\top \nabla_{\hat{\mathbf{x}}_0} \mathcal{E}) \odot \mathbf{M}
\end{equation}

\subsection{Geometric Interpretation: Vector Field Warping}
\label{subsec:geometry}

The AFS update does not merely jump between latent states. Instead, it performs \textit{Vector Field Warping}. 

\begin{figure}[t]
    \centering
    \includegraphics[width=1\linewidth]{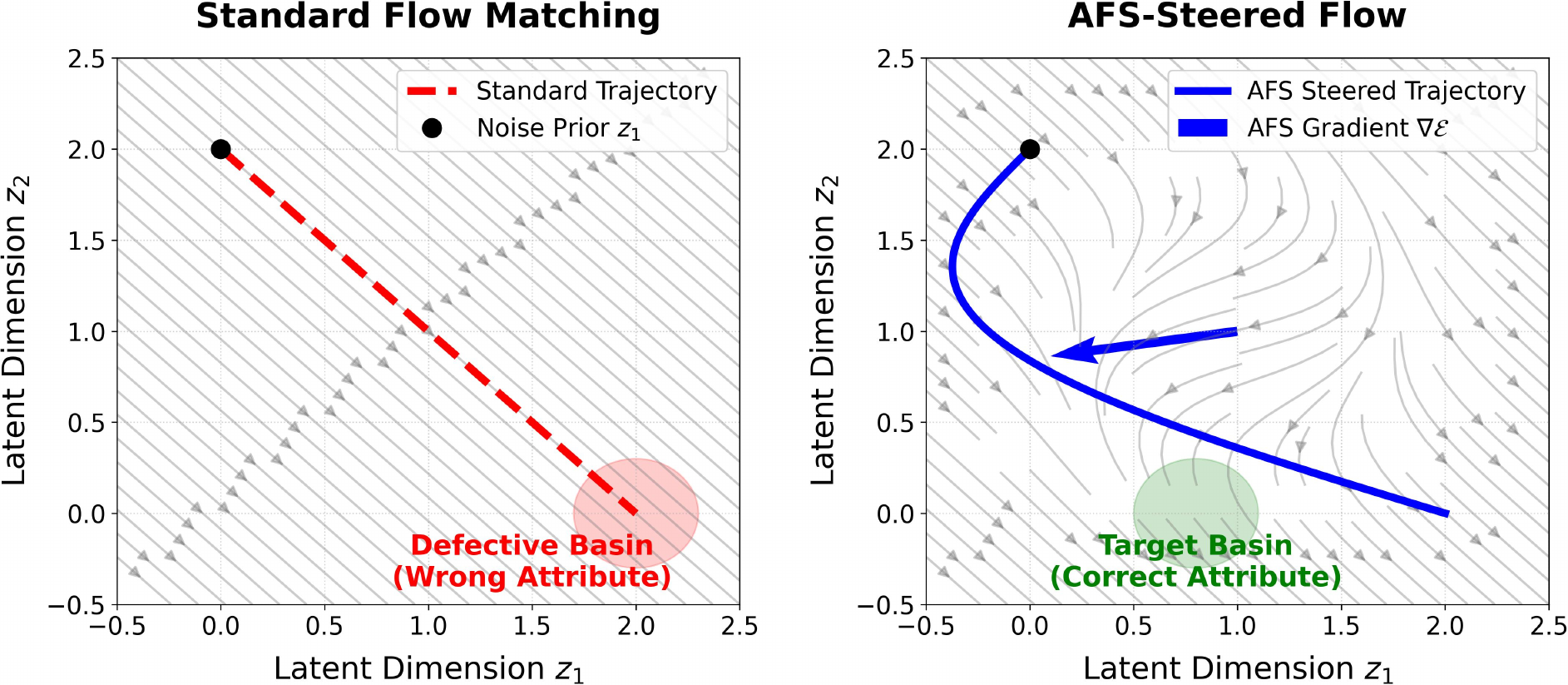}
    \caption{\textbf{Conceptual Illustration of Vector Field Warping}. (Left) Standard Flow: The trajectory follows a straight line but ends in a region with incorrect attributes (e.g., wrong color). (Right) AFS-Steered Flow: The velocity field is locally warped by the semantic gradient. The trajectory is bent toward the target basin while maintaining the overall linear transport properties of the flow.}
    \label{fig:vector_field}
\end{figure}

By modifying $\mathbf{v}_t$, we are essentially changing the momentum of the generation. This soft intervention is significantly more stable than hard latent modifications, as it preserves the cumulative ODE integration history while steering future evolution.

\subsection{Extended Hyperparameter Sensitivity Analysis}

To further investigate the robustness and controllable generation capabilities of \textbf{AFS-Search}, we conducted extensive ablation studies on three core hyperparameters: Step Size ($\eta$), Guidance Scale ($\sigma$), and Search Width ($W$). All experiments were evaluated using Qwen-VL-Max on the CompBench subset (Complex instructions) with single-round intervention.

\subsection{Quantitative Results}

\begin{figure}[t]
    \centering
    \includegraphics[width=1\linewidth]{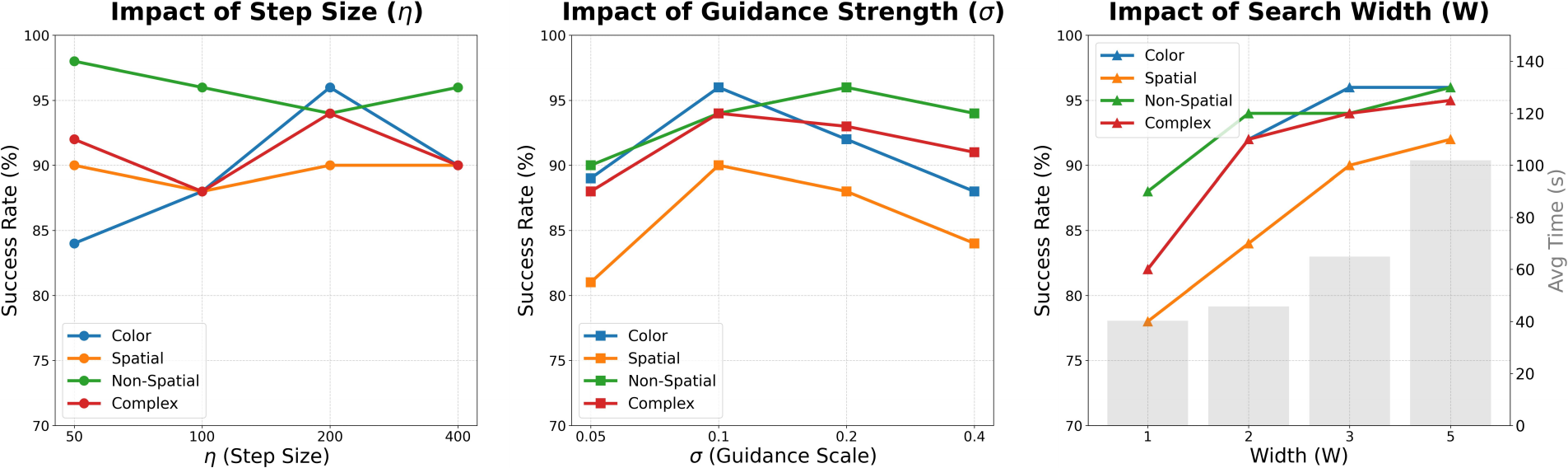}
    \caption{\textbf{Extra experiments on hyperparameters}. The success rate is calculated by Qwen-vl-max, and the scoring mechanism is provided in above prompt.}
    \label{fig:extra}
\end{figure}

The impact of these hyperparameters on Success Rate (\%) across different semantic dimensions is summarized in Fig.~\ref{fig:extra}.

\begin{itemize}
    \item \textbf{Optimal Momentum ($\eta$):} The step size $\eta$ governs the magnitude of gradient-based intervention in the latent space. We observe that $\eta=200$ serves as the ``sweet spot.'' A smaller $\eta$ ($50$) fails to provide sufficient momentum to move the latents out of local semantic minima within the limited denoising steps. Conversely, an excessive $\eta$ ($400$) introduces instability into the diffusion ODE, occasionally leading to over-correction.
    
    \item \textbf{Guidance Fidelity Trade-off ($\sigma$):} The scale parameter $\sigma$ balances the VLM-guided gradient signal with the original diffusion prior. At $\sigma=0.1$, the agent effectively corrects attribute errors without compromising image realism. At higher values ($\sigma \ge 0.2$), while the agent remains semantically focused, we observe a slight decline in Spatial scores, suggesting that aggressive guidance can sometimes disrupt the precise geometric layout.
    
    \item \textbf{Search-Performance Pareto Frontier ($W$):} Increasing the search width $W$ yields consistent performance gains. By branching the generation process, AFS-Search explores multiple denoising trajectories, significantly mitigating the risk of ``hallucinatory'' generation. However, this comes with a linear increase in computational cost. $W=3$ provides an optimal balance, achieving a $>10\%$ improvement over the greedy baseline ($W=1$) while maintaining reasonable inference latency ($\approx 65s$).
\end{itemize}

\subsection{Conclusion}
The empirical results validate the robustness of AFS-Search. The agent demonstrates high sensitivity to the search width, confirming that its MCTS-inspired exploration strategy is key to solving complex T2I tasks. For optimal performance, we recommend the configuration $\{\eta=200, \sigma=0.1, W=3\}$.

\section{Limitations and Future Works}

While our AFS-Search builds a new closed-loop paradigm in T2I field, there are still some limitations to be solved.

\noindent \textbf{Slow Inference Speed.} The agentic architecture is naturally slower in inference time compared to traditional generative models, with most of the time spent on the VLM's thinking time, the inference time of the native model, and some tool usage time. Since our paradigm successfully builds the entire framework into a self-evolving closed-loop system, future frameworks can consider how to more deeply integrate LLMs or VLMs into the native T2I model. This can not only speed up the model's inference time but also give the T2I model a deeper semantic understanding capability.

\noindent \textbf{Deeper Research in Latent Space.} For AFS-Search, the step of decoding to generate intermediate images in the middle also takes up a considerable amount of inference time. In the future, research and understanding of VLMs regarding the intermediate latent space in Flow Matching or Diffusion Models should be strengthened. If VLMs can have a better understanding of the intermediate latent space, then they can better guide the native models.

\noindent \textbf{Deeper Research in Flow Steering.} In general, the way to perform Flow steering under a native model is either a search strategy under a training-free framework or using reinforcement learning. The search strategy is simple and effective, but it can result in slower model inference and slightly higher memory usage; the reinforcement learning approach lies in how to design the reward function and quantitatively analyze the latent space within the flow-matching generation framework, an area that has been sparsely studied.

\end{document}